\documentclass[10pt,twocolumn,letterpaper]{article}

\usepackage{cvpr}
\usepackage{times}
\usepackage{epsfig}
\usepackage{graphicx}
\usepackage{amsmath}
\usepackage{amssymb}

\usepackage[pagebackref=true,breaklinks=true,letterpaper=true,colorlinks,bookmarks=false]{hyperref}
\usepackage[pagebackref=true,breaklinks=true,letterpaper=true,colorlinks,bookmarks=false]{hyperref}
\usepackage{graphicx}
\usepackage{caption}
\usepackage{multirow}
\usepackage{subfig}
\usepackage{array}
\cvprfinalcopy 

\def\cvprPaperID{1255} 

\newcommand{\Figref}[1]{Figure~\ref{fig:#1}}
\newcommand{\Tabref}[1]{Table~\ref{tab:#1}}

\newcommand{\Secref}[1]{Section~\ref{sec:#1}}
\newcommand{\figref}[1]{Fig.~\ref{fig:#1}}

\newcommand{\comment}[1]{{}}

\newcommand{\specialcell}[2][c]{%
\begin{tabular}[#1]{@{}c@{}}#2\end{tabular}}

\newcommand{\rev}[1]{{\textcolor{black}{#1}}}
\ifcvprfinal\fi
\begin{document}

\title{Deep Saliency with Encoded Low level Distance Map and High Level Features}
\author{
Gayoung Lee\\
KAIST\\
{\tt\small gylee1103@gmail.com}
\and
Yu-Wing Tai\\
SenseTime Group Limited\\
{\tt\small yuwing@gmail.com}
\and
Junmo Kim\\
KAIST\\
{\tt\small junmo.kim@kaist.ac.kr}
}

\maketitle

\begin{abstract}
Recent advances in saliency detection have utilized deep learning to obtain high level features to detect salient regions in a scene. These advances have demonstrated superior results over previous works that utilize hand-crafted low level features for saliency detection. In this paper, we demonstrate that hand-crafted features can provide complementary information to enhance performance of saliency detection that utilizes only high level features. Our method utilizes both high level and low level features for saliency detection under a unified deep learning framework. The high level features are extracted using the VGG-net, and the low level features are compared with other parts of an image to form a low level distance map. The low level distance map is then encoded using a convolutional neural network(CNN) with multiple $1 \times 1$ convolutional and ReLU layers. We concatenate the encoded low level distance map and the high level features, and connect them to a fully connected neural network classifier to evaluate the saliency of a query region. Our experiments show that our method can further improve the performance of state-of-the-art deep learning-based saliency detection methods.\end{abstract}

\section{Introduction}
Saliency detection aims to detect distinctive regions in an image that draw human attention. This topic has received a great deal of attention in computer vision and cognitive science because of its wide range of applications such as content-aware image cropping~\cite{appcrop} and resizing~\cite{appresize}, video summarization~\cite{appsum}, object detection~\cite{appdet}, and person re-identification~\cite{appreid}. Various papers such as DRFI~\cite{drfi}, GMR~\cite{gmr}, DSR~\cite{dsr}, RBD~\cite{rbd}, HDCT~\cite{hdct}, HS~\cite{hs} and GC~\cite{hc} utilize low level features such as color, texture and location information to investigate characteristics of salient regions including objectness, boundary convexity, spatial distribution, and global contrast. The recent success of deep learning in object recognition and classification~\cite{ILSVRC15} brought to a revolution in computer vision. Inspired by the human visual system, deep learning builds hierarchical layers of visual representation to extract the high level features of an image. Using extracted high level features, several recent works~\cite{legs,mdf,mcdl} have demonstrated state-of-the-art performance in saliency detection that significantly outperform previous works that utilized only low level features.

\begin{figure}
\centering
\setlength\tabcolsep{0.6pt}
\begin{tabular}{cccccc}
\subfloat{\includegraphics[width=0.16\linewidth]{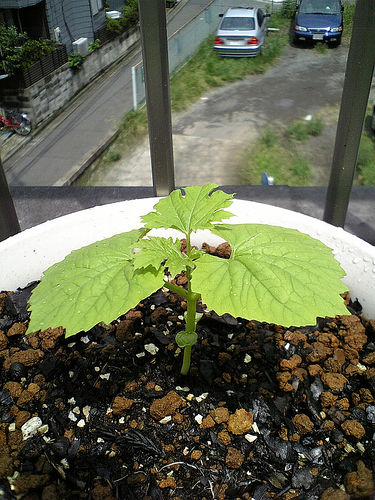}} &
\subfloat{\includegraphics[width=0.16\linewidth]{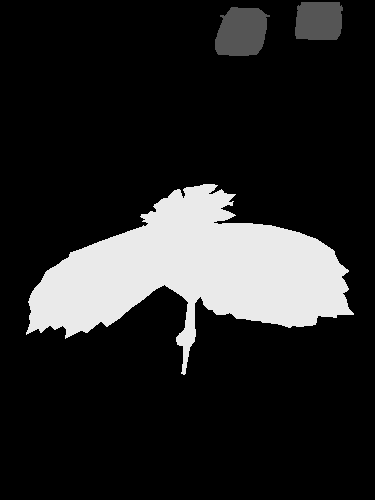}} &
\subfloat{\includegraphics[width=0.16\linewidth]{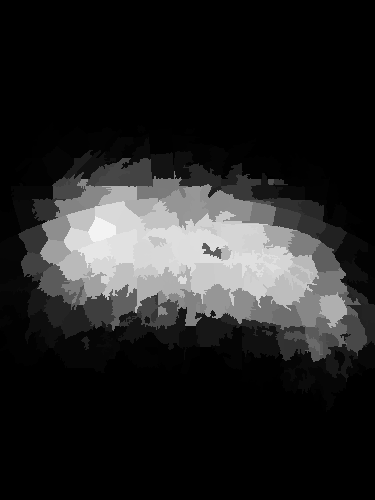}} &
\subfloat{\includegraphics[width=0.16\linewidth]{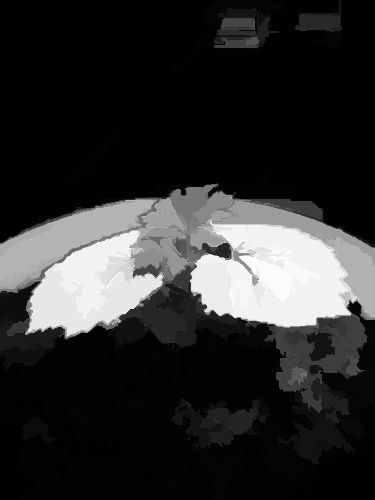}} &
\subfloat{\includegraphics[width=0.16\linewidth]{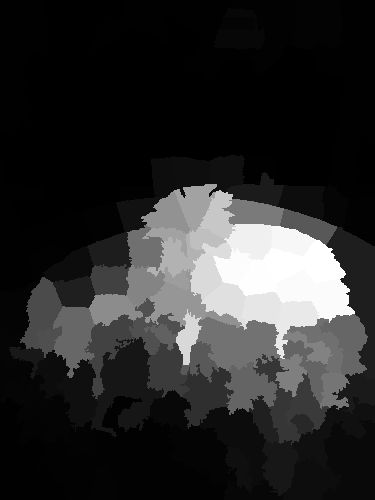}} &
\subfloat{\includegraphics[width=0.16\linewidth]{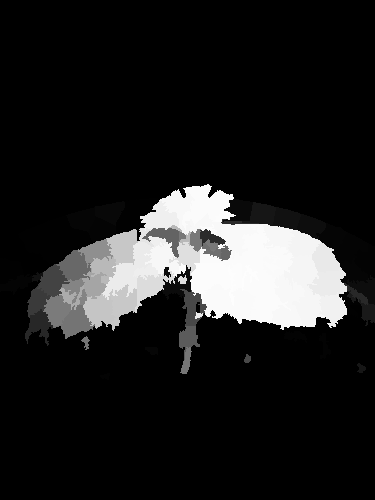}}\vspace{-0.17in}\\
\subfloat{\includegraphics[width=0.16\linewidth]{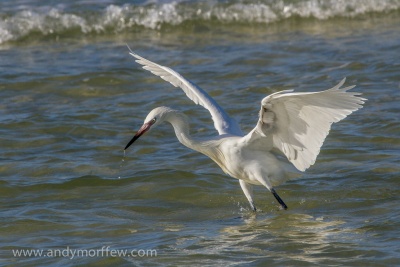}} &
\subfloat{\includegraphics[width=0.16\linewidth]{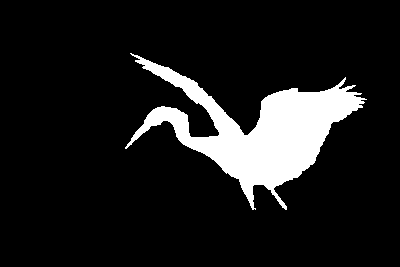}} &
\subfloat{\includegraphics[width=0.16\linewidth]{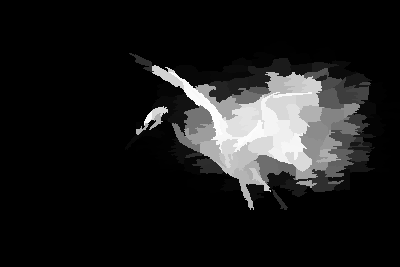}} &
\subfloat{\includegraphics[width=0.16\linewidth]{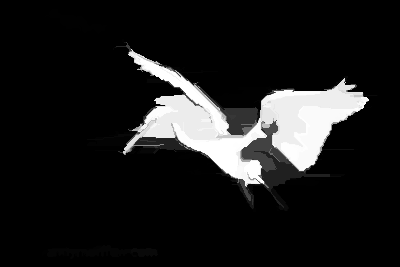}} &
\subfloat{\includegraphics[width=0.16\linewidth]{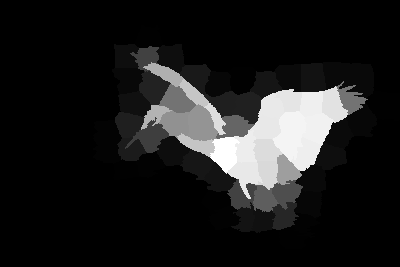}} &
\subfloat{\includegraphics[width=0.16\linewidth]{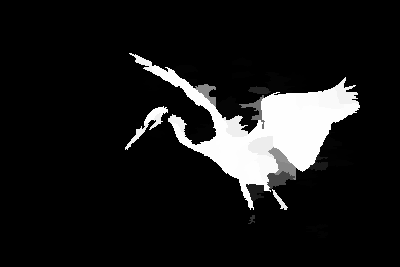}}\\
(a) & (b) & (c) & (d) & (e) & (f)\vspace{-0.10in}\\
\end{tabular}\vspace{-0.05in}
  \caption{(a) Input images, (b) Ground truth masks, (c) Fuzzy saliency masks from VGG16 features (HF setting, described in \Secref{analysis}), (d-f) Results of (d) MDF~\cite{mdf}, (e) MCDL~\cite{mcdl}, and (f) our method.}\vspace{-0.15in}
\label{fig:concept}
\end{figure}

As discussed in ~\cite{Hariharan15cvpr}, while high level features are good to evaluate objectness in an image, they are relatively weak in for determining precise localization. This is because multiple levels of convolutional and pooling layers ``blur'' the object boundaries, and high level features from the output of the last layer are too coarse spatially for the saliency detection task. This problem is illustrated in \Figref{concept}(c). To generate a precise saliency mask, previous studies utilized various methods including object proposal~\cite{legs} and superpixel classification~\cite{mdf,mcdl}. Yet, it was still very hard to differentiate salient regions from their adjacent non-salient regions because their feature distances were not directly encoded.

\begin{figure*}[t]
\centering
  \includegraphics[width=0.85\textwidth] {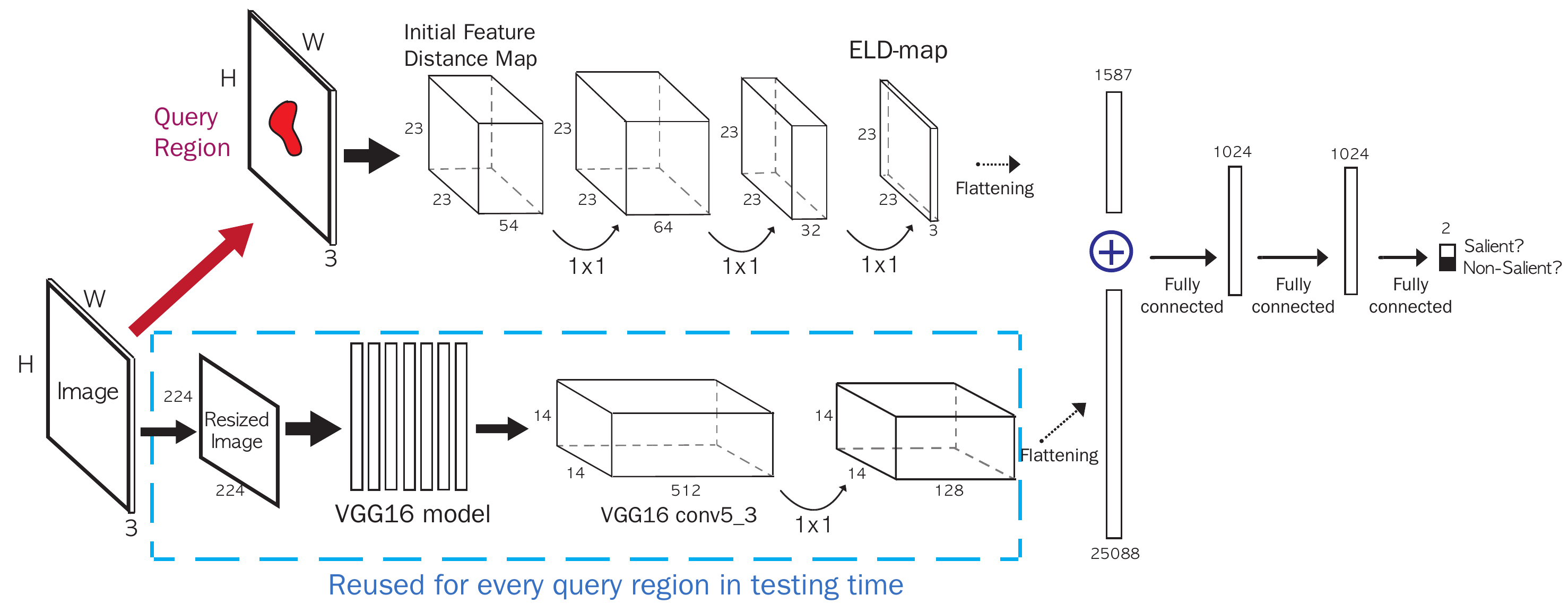}
  \vspace{-0.15in}
  \caption{Overall pipeline of our method. We compute the ELD-map from the initial feature distance map for each query region and concatenate the high level feature from the output of the conv5\_3 layer of the VGG16 model.}\vspace{-0.15in}
\label{fig:cnnarch}
\end{figure*}

In this paper, we introduce the encoded low level distance map (ELD-map), which directly encodes the feature distance between each pair of superpixels in an image. 
Our ELD-map encodes feature distance for various low level features including colors, color distributions, Gabor filter responses, and locations. Our ELD-map is unique in that it uses deep learning as an auto-encoder to encode these low level feature distances by multiple convolutional layers with $1 \times 1$ kernels. The encoded feature distance map has strong discriminative power to evaluate similarities between different parts of an image with precise boundaries among superpixels. We concatenate our ELD-map and the output of the last convolutional layer from the VGG-net (VGG16)~\cite{Simonyan14c} to form a new feature vector which is a composite of both high level and low level information. Using our new feature vector, we can precisely estimate saliency of superpixels. Without any post-processing, this method generates an accurate saliency map with precise boundaries.


In summary, our paper offers the following contributions:
\begin{itemize}
\vspace{-0.05in}
\item We introduce the ELD-map which shows that low level features can play complementary roles to assist high level features with the precise detection of salient regions.
\vspace{-0.05in}
\item Compared with previous works that utilized either high level or low level features, but not both, our work demonstrates consistent improvements across different benchmark datasets.
\vspace{-0.05in}
\item Because high level features can be reused for different query regions in an image, our method runs fast. The testing time in the ECSSD dataset~\cite{hs} takes only around 0.5 seconds per an image.
\end{itemize}

\section{Related Works}
In this section, representative works in salient region detection are reviewed. We refer readers to~\cite{SalObjSurvey} and~\cite{SalObjBenchmark} for a survey and a benchmark comparison of the state-of-the-art salient region detection algorithms.

Recent trends in salient region detection utilize learning-based approaches, which were first introduced by Liu \etal~\cite{msra10k}. Liu \etal were also the first group to released a benchmark dataset (MSRA10K) with ground truth evaluation. Following this work, several representative benchmarks with ground truth evaluation were released. These benchmarks include ECSSD~\cite{hs}, Judd~\cite{Judd09iccv}, THUR15K~\cite{thur15k}, DUTOMRON~\cite{gmr}, PASCAL-S~\cite{pascals}, and FT~\cite{Achanta09cvpr}. They cover rich variety of images containing different scenes and subjects. In addition, each one exhibits different characteristics. For example, the ground truth of the MSRA10K dataset are binary mask images which were manually segmented by human, while the ground truth of the FT~\cite{Achanta09cvpr} dataset were determined by human fixation.

Discriminative Regional Feature Integration(DRFI)~\cite{drfi}, Robust Background Detection(RBD)~\cite{rbd}, Dense and Sparse Reconstruction(DSR)~\cite{dsr}, Markov Chain(MC)~\cite{mc}, High Dimensional Color Transform(HDCT)~\cite{hdct}, and Hierarchical Saliency(HS)~\cite{hs} are the top 6 models for salient region detection reported in the benchmark paper~\cite{SalObjBenchmark}. These algorithms consider various heuristic priors such as the global contrast prior~\cite{hs} and the boundary prior~\cite{drfi} and often generate high-dimensional features to increase discriminative power~\cite{hdct, drfi} to distinguish salient regions from non-salient regions. These methods are all based on hand-crafted low level features without deep learning.

Deep learning has emerged in the field of saliency detection last year. Several methods that utilize deep learnings for saliency detection were simultaneously proposed. This includes Multiscale Deep Feature(MDF)~\cite{mdf}, Multi-Context Deep Learning(MCDL)~\cite{mcdl}, and Local Estimation and Global Search(LEGS)~\cite{legs}. They utilized high level features from the deep convolutional neural network (CNN) and demonstrated superior results over previous works that utilized only low level features. MDF and MCDL utilize superpixel algorithms, and query each region individually to assign saliency to superpixels. For each query region, MDF generates three input images that cover different scopes of an input image, and MCDL uses sliding windows with deep CNN to compute the deep features of the center superpixel. LEGS first generates an initial rough saliency mask from deep CNN and refines the saliency map using an object proposal algorithm.

Compared to the aforementioned methods, our work utilizes high level and low level features simultaneously. The high level features evaluate the objectness in an image with coarse spatial location and the low level features evaluate similarities between the different superpixels in an image. Our high level and low level features are combined and evaluated by a multi-level fully connected neural network classifier, that seamlessly considers both high level and low level features to assign saliency to query superpixels. Experiments demonstrate that our method significantly outperforms previous methods that utilize either low level features or high level features, but not both.

\section{Algorithms}
The overall pipeline of our method is illustrated in \Figref{cnnarch}. First, the process for construction of the ELD-map is described. Then, we describe how the high level features were extracted and integrated with the ELD-map for salient region classification. At the end of this section, we report the results of our self evaluations to analyze the effects of the ELD-map and the high level features in our saliency detection framework.

\subsection{Construction of the ELD-map}
\label{sec:eld}
\begin{figure}[t]
\centering
  \includegraphics[width=0.45\textwidth]{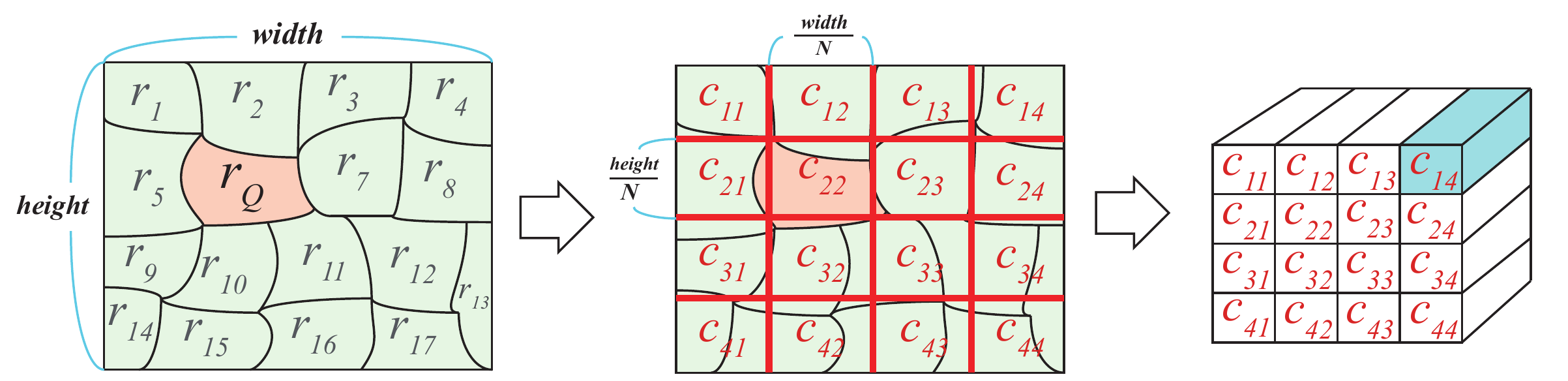}\vspace{-0.05in}
  \caption{Visualization of the construction process for the initial low level feature distance map. Each grid cell, which represents uniformly divided area of an image, is described by the features of the superpixel that occupies the largest area of the grid cell. Using the features, we construct an $N \times N \times K$ feature distance map. The computed features and distances are summarized in \Tabref{regional_descriptor} and \Tabref{cell_descriptor} } \vspace{-0.15in}
  \label{fig:initialmap}
\end{figure}

\begin{table}
\begin{center}
\begin{tabular}{|c|c|}
\hline
Features of a superpixel ($f(r_{c})$) & \specialcell{Feature Index} \\
\hline\hline
Average RGB value & 1-3 \\
Average LAB value & 4-6 \\
Average HSV value & 7-9 \\
Gabor filter response & 10-33\\
Maximum Gabor response & 34\\
Center location & 35-36\\
RGB color histogram & 37-61\\
LAB color histogram & 62-86\\
HSV color histogram & 87-110\\
\hline
\end{tabular}
\end{center}
\vspace{-0.15in}
\caption{The list of extracted features of a superpixel.}
\vspace{-0.1in}
\label{tab:regional_descriptor}
\begin{center}
\begin{tabular}{|c|c|c|}
\hline
Distance map features & \specialcell{\#$f(\cdot)$} & \specialcell{Feature Index}\\
\hline\hline
  $f(c_{ij}) - f(r_q)$ & 1-36 & 1-36\\
\hline
\specialcell{$\chi^2$ distance($f(c_{ij}), f(r_q)$)} & 37-110 & 37-45\\
\hline
  $f(c_{ij})$ & 1-9 & 46-54\\
\hline
\end{tabular}
\end{center}
\vspace{-0.15in}
\caption{The list of feature distances used for computing the initial low level feature distance map. $f(r_q)$ is the extracted features of a query superpixel, $r_q$, and
$f(c_{ij})$ is the extracted features of a grid cell $c_{ij}$, where $f(c_{ij}) := f(r_c^*)$. Details are described in \Secref{eld}.}\vspace{-0.1in}
\label{tab:cell_descriptor}
\end{table}

Our algorithm utilizes a superpixel-based approach for saliency detection. To segment an image into superpixels, the SLIC\cite{slic} algorithm is used. The major benefits of using the SLIC algorithm for superpixel segmentation are that the segmented superpixels are roughly regular and that it provides control on the number of superpixels.

After superpixel segmentation, the initial hand-crafted low level features of each superpixel are calculated, and the superpixel representation is converted into a regular grid representation as illustrated in \Figref{initialmap}. To be more specific, we assign superpixels to grid cells according to their occupying area in each cell. This regular grid representation is efficient for CNN architecture because we can convert images with different resolutions and aspect ratios into a fixed size distance map without resizing and cropping. 

In our implementation, the size of the regular grid was set to $23\times23$. We index the superpixels as $S = \{r_1,...,r_M\}$, and the grid cells of the regular grid as $G = \{c_{11},c_{12},...,c_{NN}\}$, $N=23$. We denote the computed feature descriptor of each superpixel region as $f(r_c)$. The collected features for each superpixel are summarized in \Tabref{regional_descriptor}. Our hand-crafted features are all low level features related to colors (average colors in RGB, LAB, and HSV spaces, and their local color histograms), textures (Gabor filter responses~\cite{gabor} averaged over pixels in each region), and locations (center location of a superpixel). We normalize the pixel coordinates so that the range of coordinates was within $[0,1]$ and include the maximum over 24 values for the Gabor filter response in each region. Each grid cell descriptor is equal to the descriptor of the superpixel which occupies the largest area inside that grid cell, \ie, $f(c_{ij}) := f(r_c^*)$, where $r_c^* = \arg\max_{r_c} { \#pixels({r_c} \cap c_{ij})} $.

Similar to MCDL\cite{mcdl} and MDF\cite{mdf}, we query the saliency score of each region individually. For each query region, we compute a low level feature distance map that modelled the feature distances between the queried superpixel $f(r_q)$ and grid cells $f(c_{ij})$ in the regular grid. For the mean color value and Gabor response, we simply compute the differences within them where negative values are allowed, and use the Chi-square ($\chi^2$) distance for color histograms between $r_c^*$ and $r_q$. We attach the average colors of $f(c_{ij})$ at the end of the distance measurements as a reference point, and find that this improved the performance. \Tabref{cell_descriptor} summarizes the computed feature distances of the initial feature distance map where the number of the initial features ($K$) is 54. After computing the distances, the size of the initial feature distance map becomes $23 \times 23 \times 54$.

The initial feature distance map is then encoded to a compact but accurate feature distance map using the multiple $1 \times 1$ convolutional and ReLU layers, as illustrated in \Figref{cnnarch}. The multiple $1 \times 1$ convolutional and ReLU layers work as a fully connected layer across channels to find the best nonlinear combination of feature distances that better describe the similarities and dissimilarities between a query superpixel and the other regions of an image. Because the dimension of the initial map is reduced, we call this distance map as an encoded low level distance map (ELD-map). In our implementation, the size of the ELD-map was $23 \times 23 \times 3$. In the self-evaluation experiment in \Tabref{controlled_table}, we find that encoding the low level feature distance map with the deep CNN with $1 \times 1$ kernel enhances the performance of our method. The effects of the encoding will be discussed in \Secref{analysis}.

\subsection{Integration with High Level Features}
We extract the high level features using the VGG16 model pretrained by the ImageNet Dataset~\cite{ILSVRC15}. The \textbf{VGG16}~\cite{Simonyan14c} won the ImageNet2014 CLS-LOC task. We used the VGG16 model distributed by Caffe Model Zoo~\cite{caffe} \rev{without finetuning}. We resize the input images to $224\times 224$ to fit to the fixed input size of the VGG16 model and extract a ``conv5\_3'' feature map, which is generated after passing the last convolutional layer. The extracted features has 512 channels and $14\times 14$ resolution. To fit the features to our GPU memory, we attach an additional convolutional layer with a $1\times 1$ kernel for feature selection and dimensionality reduction as in GoogleNet~\cite{googlenet}.

For each input image, we process it with the pre-trained deep CNN only once and reuse the extracted high level feature map for all queried regions. Therefore, our computational cost is small even when we use a very deep and powerful model such as the VGG16 model. Although other parts of our algorithm, including generating the ELD-map and applying fully-connected layers, should be repeated each time, the cost from these parts is much smaller than running the VGG16 model.

Before applying the fully-connected layers to classify the queried region, we concatenate the ELD-map and ``conv5\_3'' feature map after flattening each map. Afterwards, two fully-connected layers with 1024 nodes generate a saliency score for the queried region using the concatenated features. We use the cross entropy loss for softmax classifier to evaluate the outputs:
\begin{equation}\label{eq:softmax}
L = -\sum_{j=0}^{1}1_{(y=j)}\log(\frac{e^{z_j}}{e^{z_0} + e^{z_1}})
\end{equation}
where $0$ and $1$ denote non-salient and salient region labels respectively, and $z_0$ and $z_1$ are the score of each label of training data. Since the ELD-map features and the high level features are fixed in length, their spatial correlation can be learnt from training data automatically in the fully connected layers.

\subsection{Analysis of the Effects of the Encoded Low level Distance map}
\label{sec:analysis}

\begin{figure}
\centering
\setlength\tabcolsep{1pt}
\begin{tabular}{cccccc}
\subfloat{\includegraphics[width=0.16\linewidth]{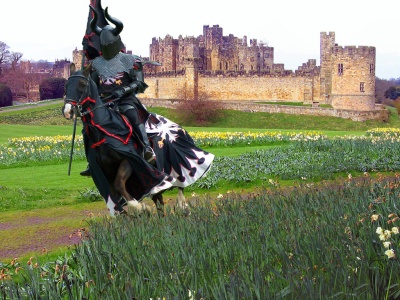}} &
\subfloat{\includegraphics[width=0.16\linewidth]{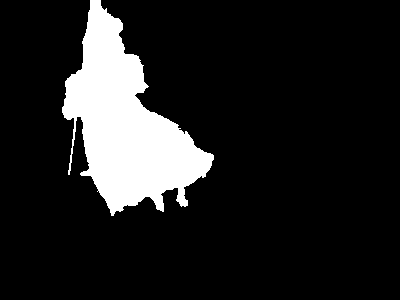}} &
\subfloat{\includegraphics[width=0.16\linewidth]{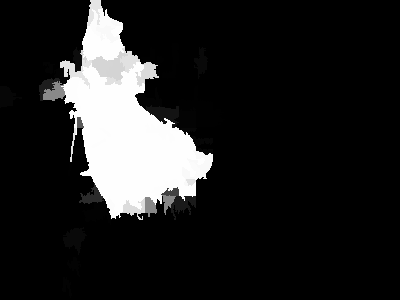}} &
\subfloat{\includegraphics[width=0.16\linewidth]{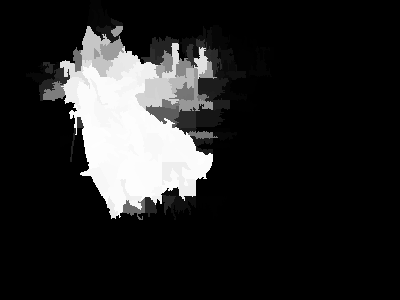}} &
\subfloat{\includegraphics[width=0.16\linewidth]{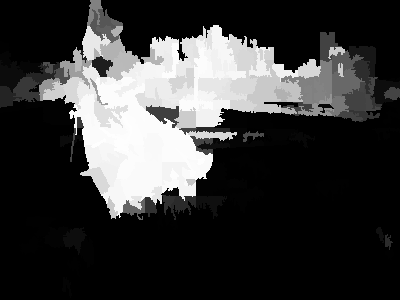}} &
\subfloat{\includegraphics[width=0.16\linewidth]{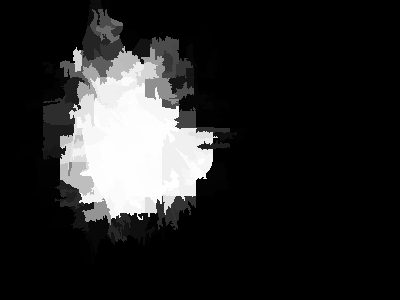}}\vspace{-0.15in}\\
\subfloat{\includegraphics[width=0.16\linewidth]{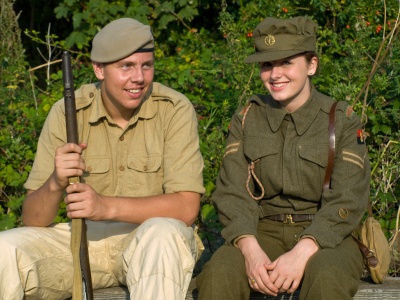}} &
\subfloat{\includegraphics[width=0.16\linewidth]{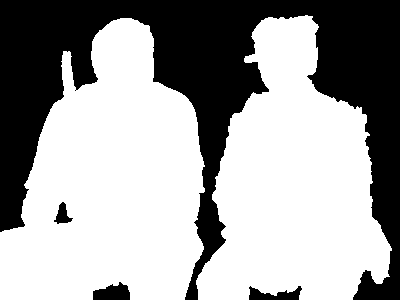}} &
\subfloat{\includegraphics[width=0.16\linewidth]{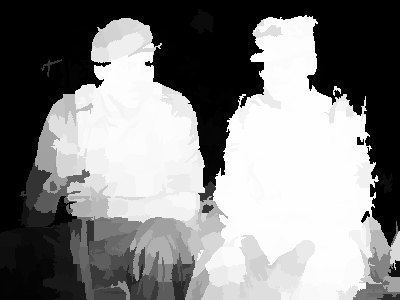}} &
\subfloat{\includegraphics[width=0.16\linewidth]{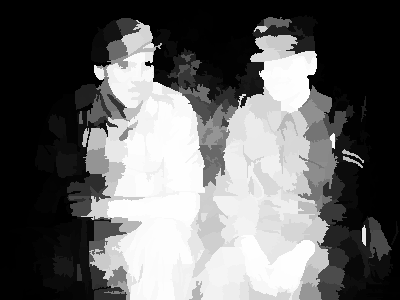}} &
\subfloat{\includegraphics[width=0.16\linewidth]{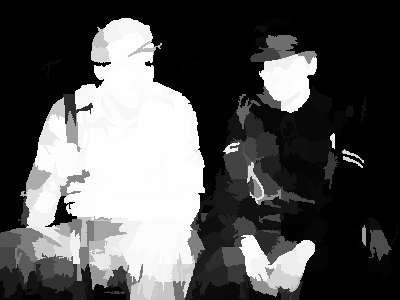}} &
\subfloat{\includegraphics[width=0.16\linewidth]{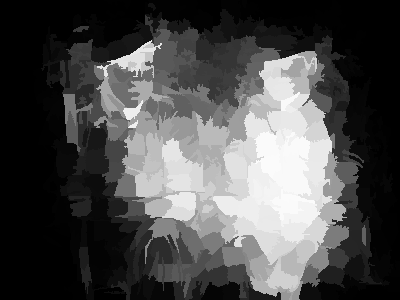}}\vspace{-0.15in}\\
\subfloat{\includegraphics[width=0.16\linewidth]{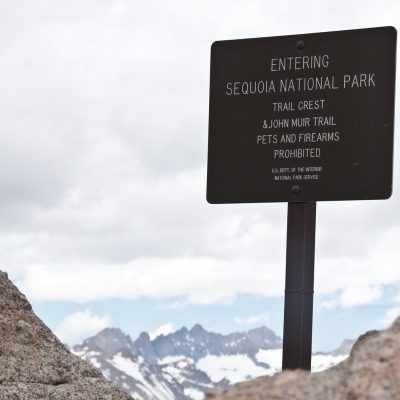}} &
\subfloat{\includegraphics[width=0.16\linewidth]{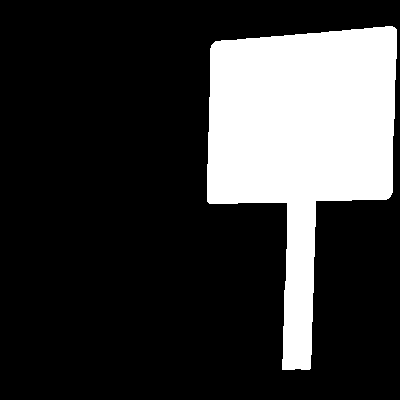}} &
\subfloat{\includegraphics[width=0.16\linewidth]{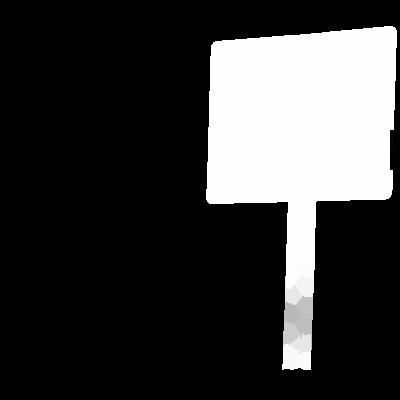}} &
\subfloat{\includegraphics[width=0.16\linewidth]{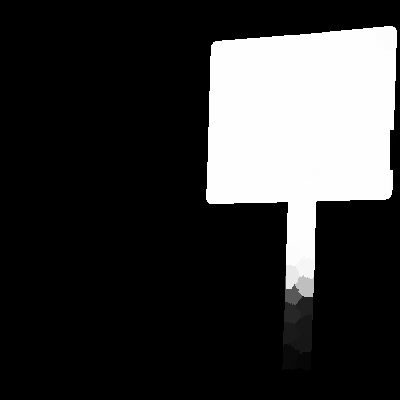}} &
\subfloat{\includegraphics[width=0.16\linewidth]{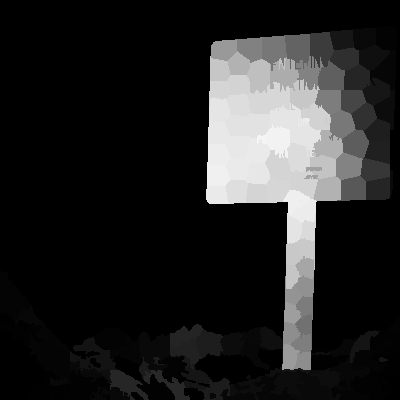}} &
\subfloat{\includegraphics[width=0.16\linewidth]{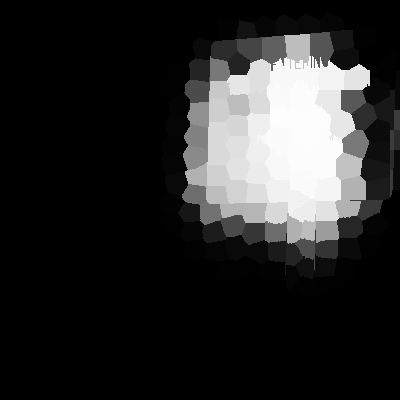}}\\
(a) & (b) & (c) & (d) & (e) & (f) \\
\end{tabular}\vspace{-0.1in}
  \caption{Visual comparisons of results in our self-evaluation experiments. (a) Input images, (b) Ground truth masks, (c-f) the results of our algorithm (c) using both ELD-map and high level features (ELD-HF) (d) using both non-encoded low level distance map and high level features (LD-HF) (e) using only encoded low level distance map (ELD) (f) using only high level features (HF). Details of each experiment are described in \Tabref{controlled_table}.}\vspace{-0.15in}
\label{fig:controlled}
\end{figure}

\begin{table*}
\begin{center}
\begin{tabular}{|c|c|c|c|c|c|}
\hline
\specialcell{Setting\\Description} & \specialcell{Encoded Low level \\Distance map} & \specialcell{Non-encoded Low level \\Distance map} & \specialcell{High level features\\from VGG16} 
& \specialcell{f-measure on\\ECSSD}& \specialcell{f-measure on\\PASCAL-S}\\
\hline
ELD-HF & Use & Not Use & Use & 0.867 & 
0.770\\
\hline
LD-HF & Not Use & Use &
Use & 0.835 &
0.735\\
\hline
ELD & Use & Not Use &
Not Use & 0.790 &
0.682\\
\hline
HF & Not Use & Location Only & 
Use & 0.768 &
0.693\\
\hline
\end{tabular}
\end{center}
\vspace{-0.15in}
\caption{The detail of settings of the controlled experiments. Using both ELD-map and high level features from VGG16 shows the best performance.}\vspace{-0.15in}
\label{tab:controlled_table}
\end{table*}

\begin{figure}
\centering
\setlength\tabcolsep{-0.25in}
\vspace{-0.1in} 
\begin{tabular}{cc}
\subfloat{\includegraphics[width=0.67\linewidth]{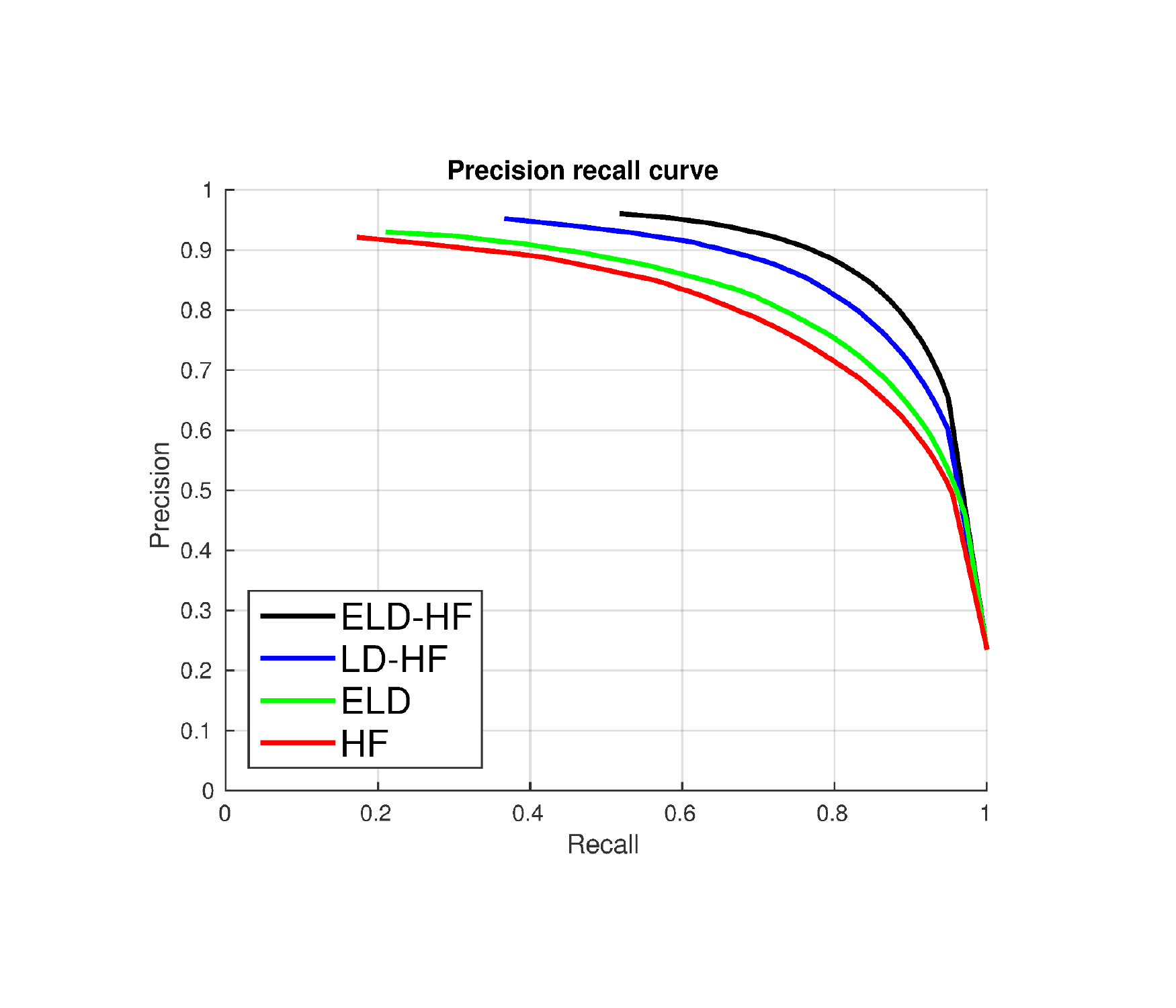}} &
\subfloat{\includegraphics[width=0.67\linewidth]{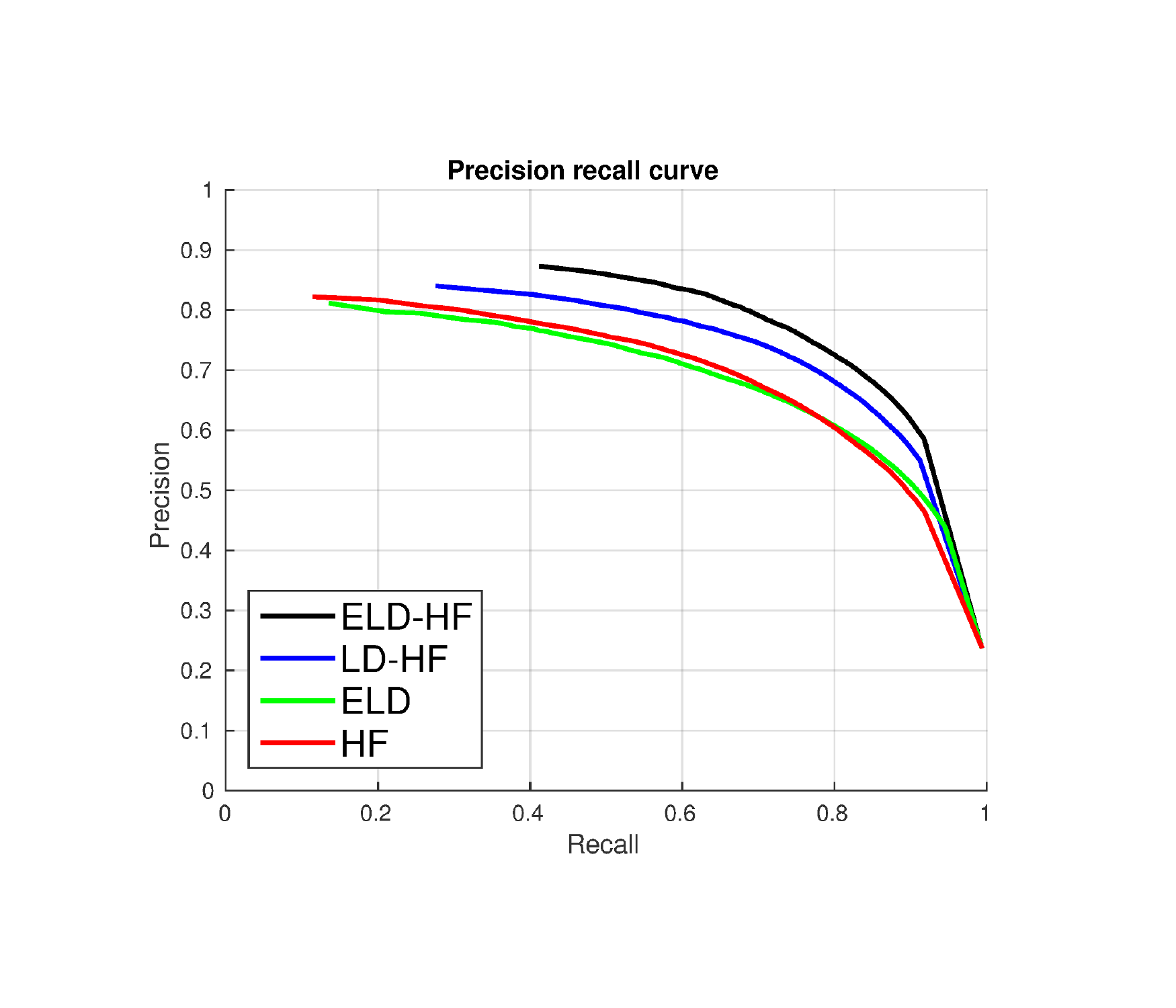}}\vspace{-0.2in} \\
(a) ECSSD  & (b) PASCAL-S\\
\end{tabular}\vspace{-0.1in}
\caption{Precision-Recall graphs of the controlled experiments described in \Tabref{controlled_table} }\vspace{-0.1in}\label{fig:controlled_prgraph}
\end{figure}

\begin{figure}
\centering
\setlength\tabcolsep{-0.25in}
\vspace{-0.4in} 
\begin{tabular}{cc}
\subfloat{\includegraphics[width=0.67\linewidth]{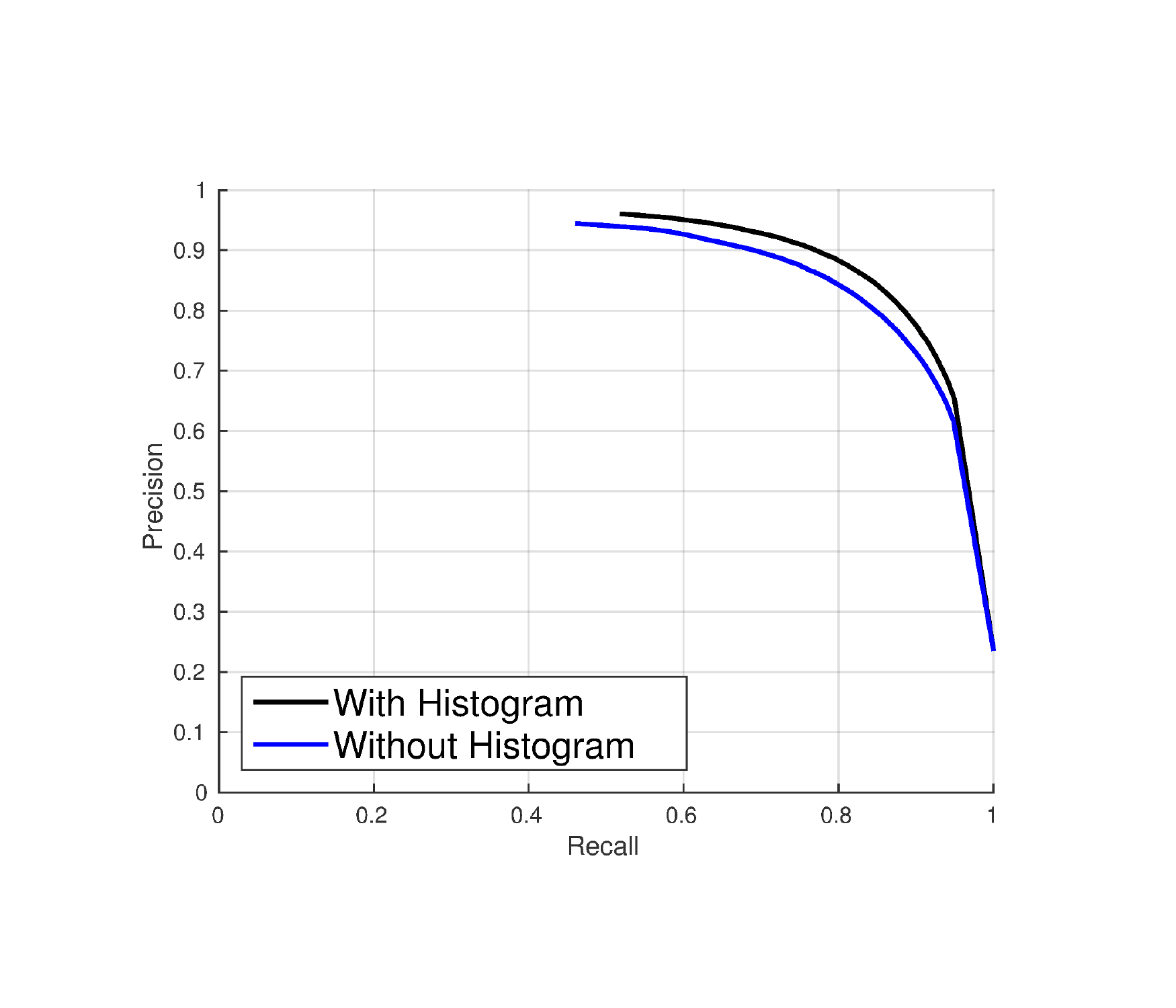}} &
\subfloat{\includegraphics[width=0.67\linewidth]{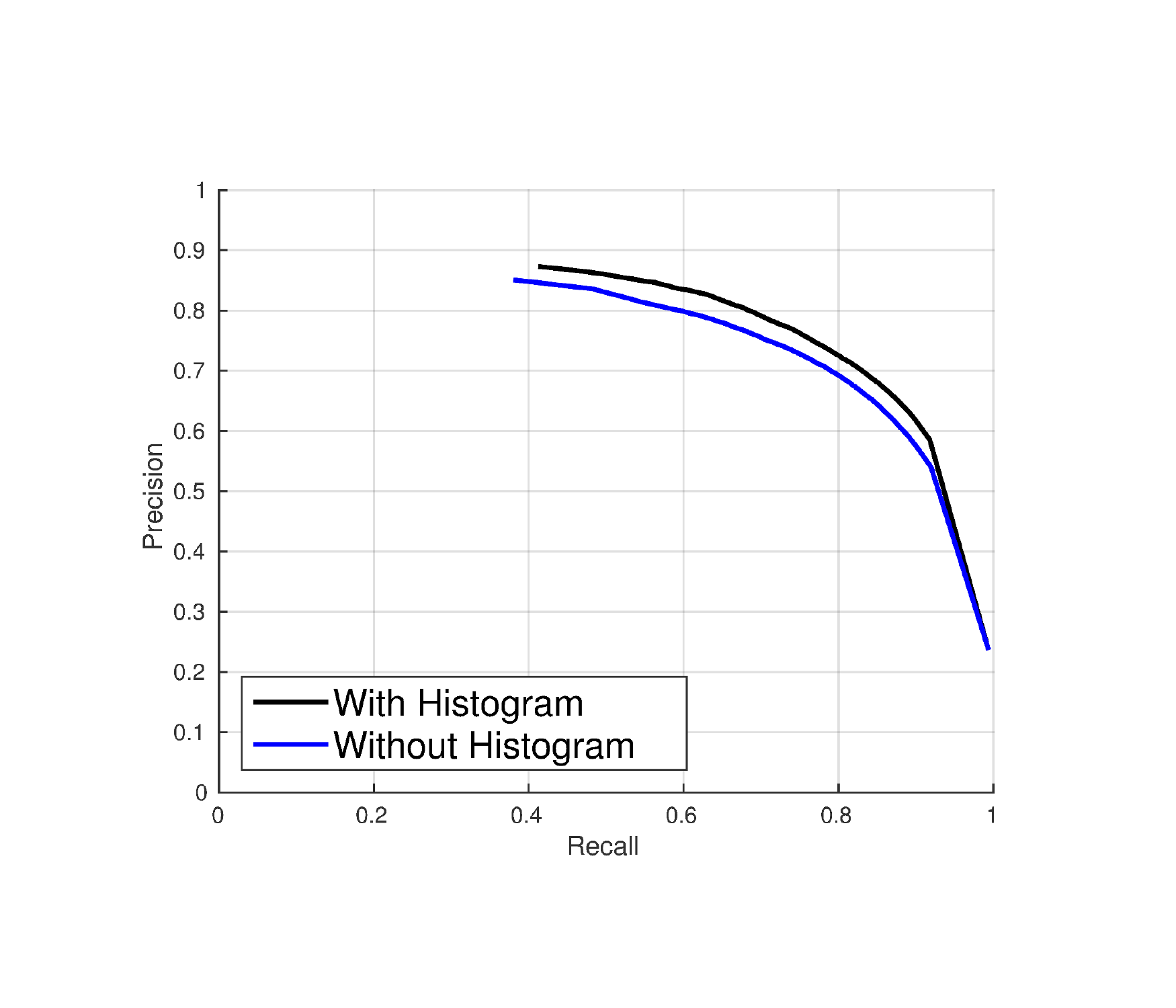}}\vspace{-0.2in}\\
(a) ECSSD  & (b) PASCAL-S\\
\end{tabular}\vspace{-0.1in}
\caption{Precision-Recall graphs of the controlled experiments to show the effect of the statistical features. }\vspace{-0.1in}\label{fig:controlled_histo}
\end{figure}

\begin{figure*}
\centering
\setlength\tabcolsep{1pt}
\begin{tabular}{cccccccccc}
\subfloat{\includegraphics[width=0.092\textwidth]{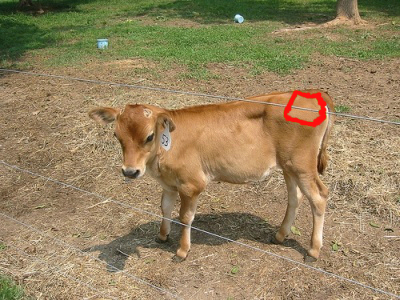}} &
\subfloat{\includegraphics[width=0.092\textwidth]{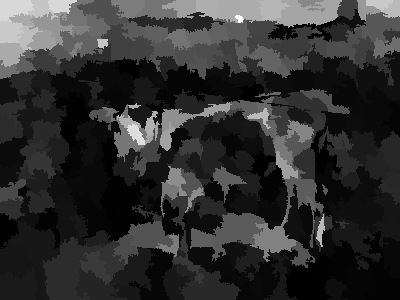}} &
\subfloat{\includegraphics[width=0.092\textwidth]{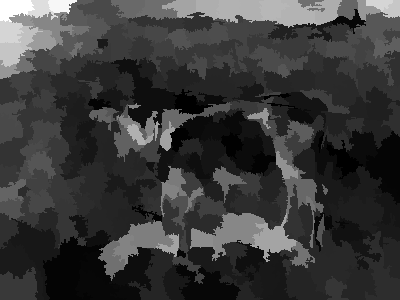}} &
\subfloat{\includegraphics[width=0.092\textwidth]{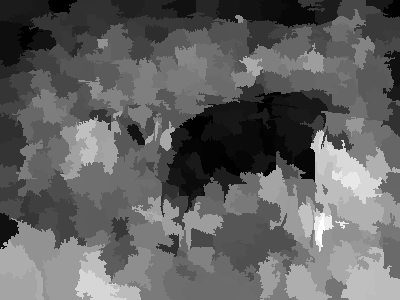}} &
\subfloat{\includegraphics[width=0.092\textwidth]{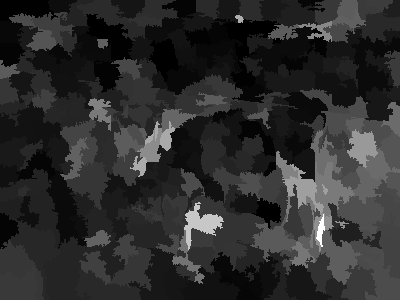}} &
\subfloat{\includegraphics[width=0.092\textwidth]{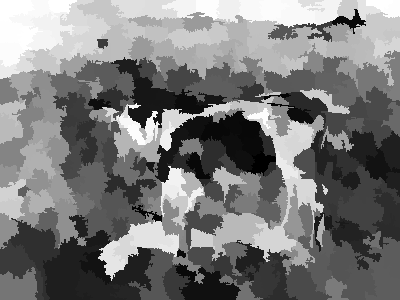}} &
\subfloat{\includegraphics[width=0.092\textwidth]{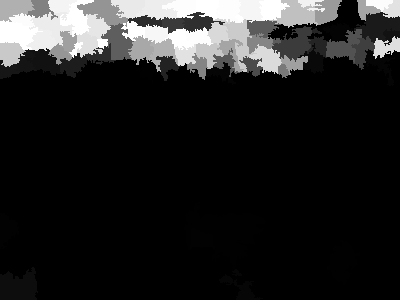}} &
\subfloat{\includegraphics[width=0.092\textwidth]{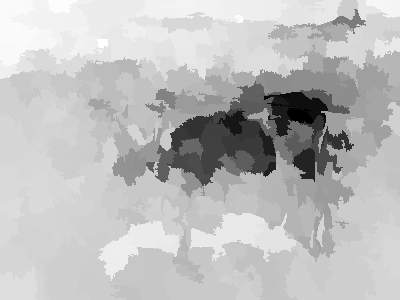}} &
\subfloat{\includegraphics[width=0.092\textwidth]{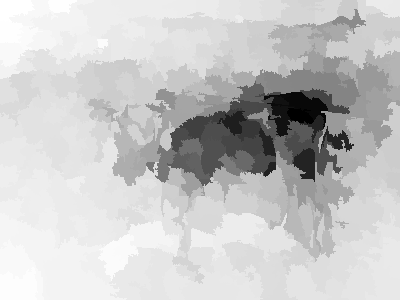}} &
\subfloat{\includegraphics[width=0.092\textwidth]{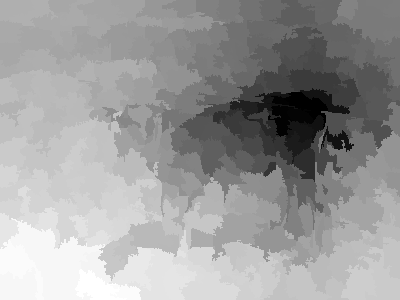}}\vspace{-0.15in}\\
\subfloat{\includegraphics[width=0.092\textwidth]{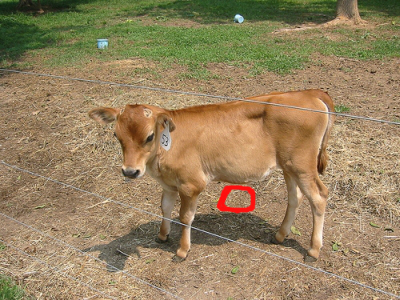}} &
\subfloat{\includegraphics[width=0.092\textwidth]{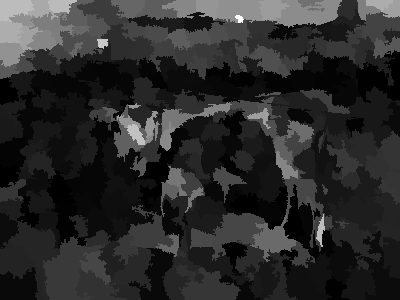}} &
\subfloat{\includegraphics[width=0.092\textwidth]{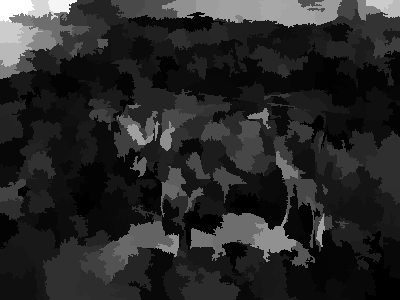}} &
\subfloat{\includegraphics[width=0.092\textwidth]{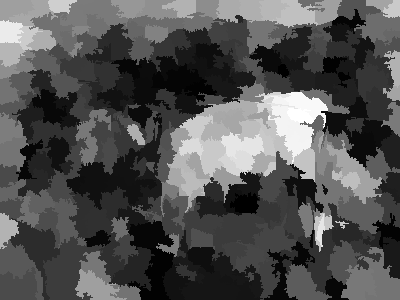}} &
\subfloat{\includegraphics[width=0.092\textwidth]{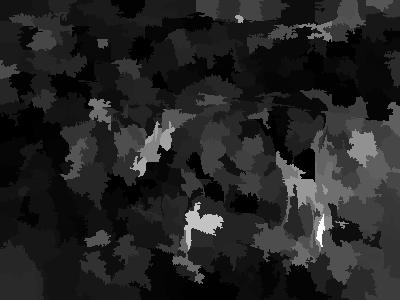}} &
\subfloat{\includegraphics[width=0.092\textwidth]{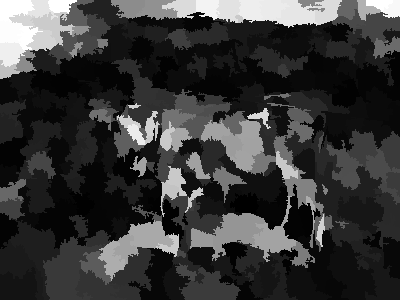}} &
\subfloat{\includegraphics[width=0.092\textwidth]{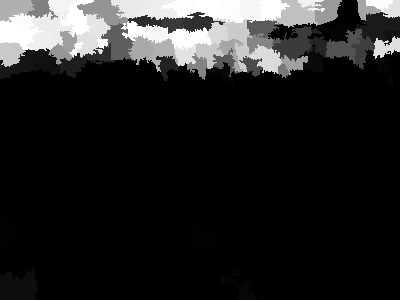}} &
\subfloat{\includegraphics[width=0.092\textwidth]{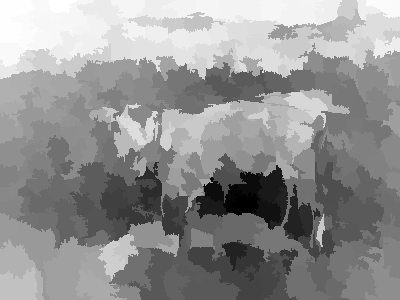}} &
\subfloat{\includegraphics[width=0.092\textwidth]{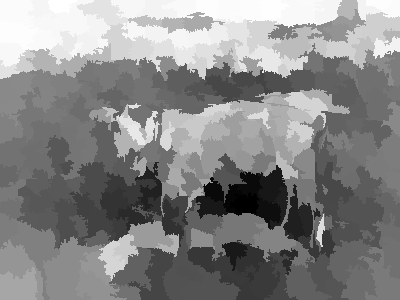}} &
\subfloat{\includegraphics[width=0.092\textwidth]{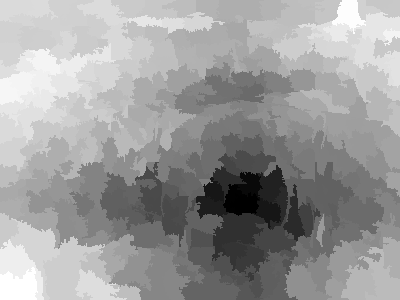}}\vspace{-0.15in}\\
\subfloat{\includegraphics[width=0.092\textwidth]{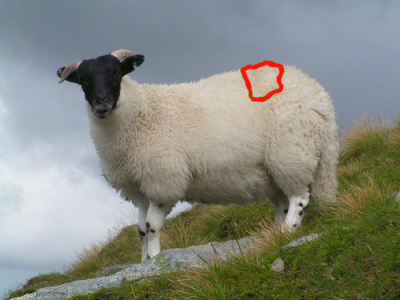}} &
\subfloat{\includegraphics[width=0.092\textwidth]{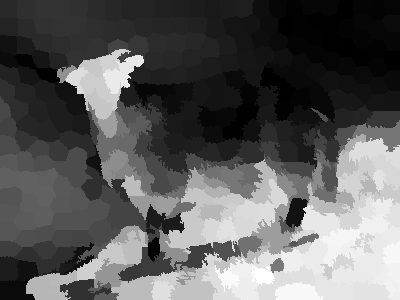}} &
\subfloat{\includegraphics[width=0.092\textwidth]{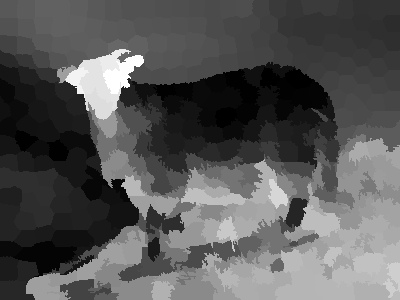}} &
\subfloat{\includegraphics[width=0.092\textwidth]{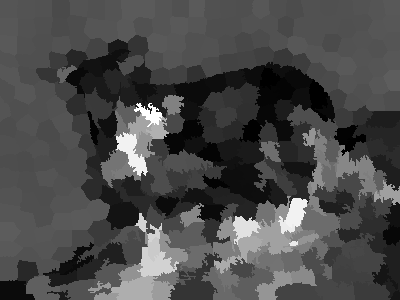}} &
\subfloat{\includegraphics[width=0.092\textwidth]{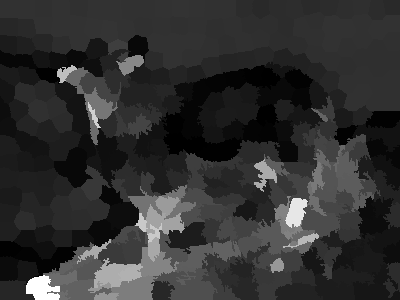}} &
\subfloat{\includegraphics[width=0.092\textwidth]{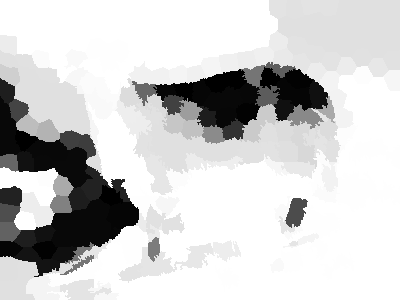}} &
\subfloat{\includegraphics[width=0.092\textwidth]{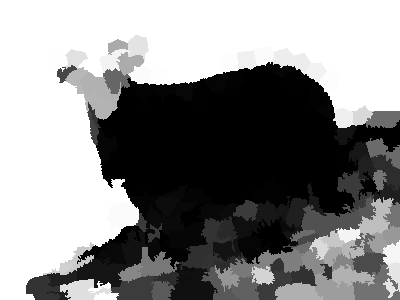}} &
\subfloat{\includegraphics[width=0.092\textwidth]{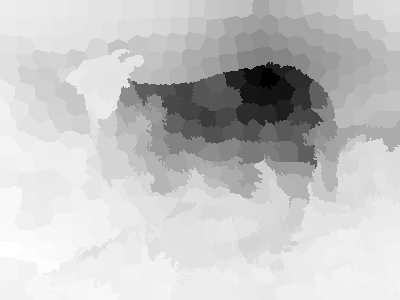}} &
\subfloat{\includegraphics[width=0.092\textwidth]{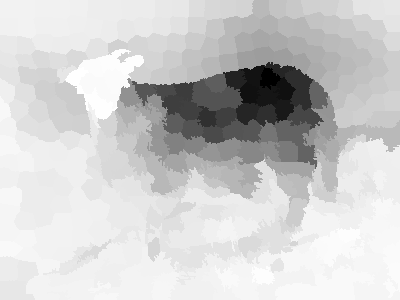}} &
\subfloat{\includegraphics[width=0.092\textwidth]{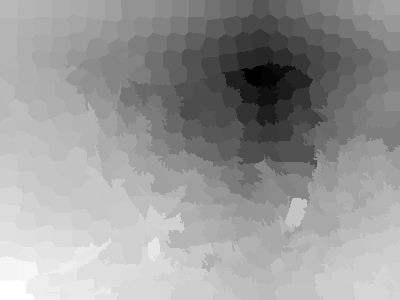}}\\
(a) & (b) & (c) & (d) & (e) & (f) & (g) & (h) & (i) & (j)\\
\end{tabular}
\vspace{-0.1in}
  \caption{Comparisons of the discriminative power of different features and our ELD-map feature space. (a) Input images, the query superpixels are highlighted. (b)-(g) are the distance maps of the different features between the query superpixel and other superpixels in an image. (b)-(c) Distance maps of average color of (b) R-channel (RGB color space), and (c) L-channel (LAB color space). (d) Differences of the first Gabor filter responses. (e) Differences of the maximum gabor filter responses. (f)-(g) Chi-square distance maps of (f) L-channel histogram (LAB color space), and (g) H-channel histogram (HSV color space).  (h)-(j) our Encoded Low level Distance map (ELD-map).}\vspace{-0.1in}
\label{fig:3ch_compare}

\end{figure*}

\rev{Although theoretically neural networks can model any complex function\cite{hornik1989multilayer}, practically they may suffer from limited training data and limited computational resources. For instance, overfitting and underfitting frequently occur because of a small dataset and the complexity of desired features. It is also common for CNN to generate feature maps with much lower resolution than original input images. By providing strongly relevant information, the encoded low level distance map(ELD-map) complements the features from deep CNN and guides the classifier to learn properly. ELD-map has two main advantages:} (1) it can easily generate the fine-grained dense saliency mask, and (2) it provides additional low level feature distances, which can be hard to learn for CNN, such as Chi-square distance between histograms.

We performed multiple controlled experiments to demonstrate the effects of the ELD-map in our algorithm. We conducted the experiments using four different settings: The \textbf{ELD-HF} setting uses both the ELD-map and the high level feature map from the VGG16 model. The \textbf{LD-HF} setting utilizes both the low level feature distances and the high level feature map, but does not encode the low level distances with the $1\times 1$ convolutional network. The \textbf{ELD} setting uses only ELD-map without high level features from the VGG16 model. The \textbf{HF} setting uses the high level feature map from VGG16 model and the location distance between the query region and other regions to notify which region is queried. We ran all models until the training data loss converged.

The results of the controlled experiments are shown in \Figref{controlled}. The model using only the high level feature map from the deep CNN detected the approximate location of the salient objects but was unable to capture detailed location because the high level feature maps had lower resolution than the resolution of the original input images. On the other hand, the model with only the low level features failed to distinguish the salient object in the first row. With both the low level feature distances and the high level feature map, the models could correctly capture salient objects and their exact boundaries. Also, we found that the ELD-map often helps to find salient objects that are difficult to detect using only CNN as shown in the second row. We speculate that the ELD-map can provide additional information which is hard to be accurately modeled by the convolutional layers. \rev{Some of the hand-crafted features of our method are statistical features, \eg histogram, and we use $\chi^2$ distance to measure the distance between histograms that would be difficult to learn by CNN. To demonstrate the effects of the statistical features, we re-train our network with the histogram features removed from our network. The comparisons are shown in \figref{controlled_histo}. Clearly, the histogram features improve the performance of our work. Similarly, for features in other color space, \eg LAB and HSV, it may require more layers to model such transformation, but we can easily adopt them from hand-crafted features.}

\Tabref{controlled_table} summarizes the controlled experiments for the self-evaluation of our method. It also shows the quantitative comparisons in terms of f-measure on the ECSSD and the PASCAL-S datasets. The corresponding quantitative comparisons in terms of the Precision-Recall graphs are presented in \Figref{controlled_prgraph}. The model utilizing both ELD-map and high level features exhibits the best performance. By comparing ELD-HF and LD-HF settings, we found that it is useful to apply $1\times 1$ kernels among the low level features.

\Figref{3ch_compare} shows the initial hand-crafted distance features and ELD-map. For the ELD-map, which is originally the $23\times 23$ size grid, we visualized each superpixel using the feature value of the closest grid cell according to the location of the center pixel. Each hand-crafted feature has its own weakness but it captures different aspects of similarities or dissimilarities between superpixels. Our $1\times 1$ kernels work as fully-connected layers among low level feature distances and generate a powerful feature distance map by combining all of the original feature distances nonlinearly. This nonlinear mapping is data-driven which is directly learnt from training data automatically. We can see the strong discriminative power of feature distances in ELD-map. While the third channel (j) is related to the position of the query region, the other two channels (h-i) seem to indicate the differences of appearance such as color and texture. Therefore, the ELD-map helps to group regions that belong to the same object, because regions which have the similar color and texture have similar values in the two channels of the ELD-map regardless of their position.


\section{Experiment and Discussion}
\begin{figure*}
\centering
\setlength\tabcolsep{-0.2in}
\begin{tabular}{ccccc}
\subfloat{\includegraphics[width=0.26\textwidth]{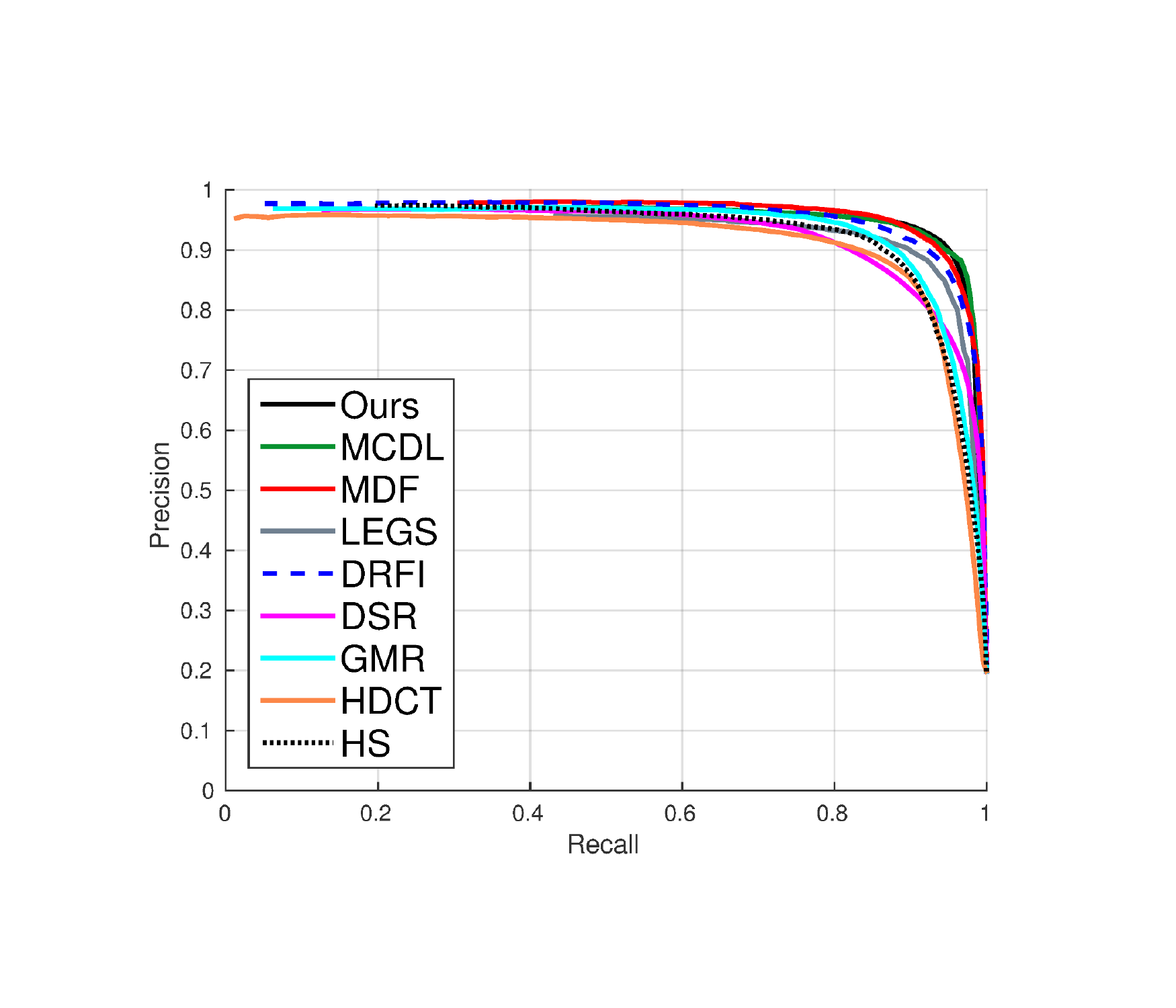}} &
\subfloat{\includegraphics[width=0.26\textwidth]{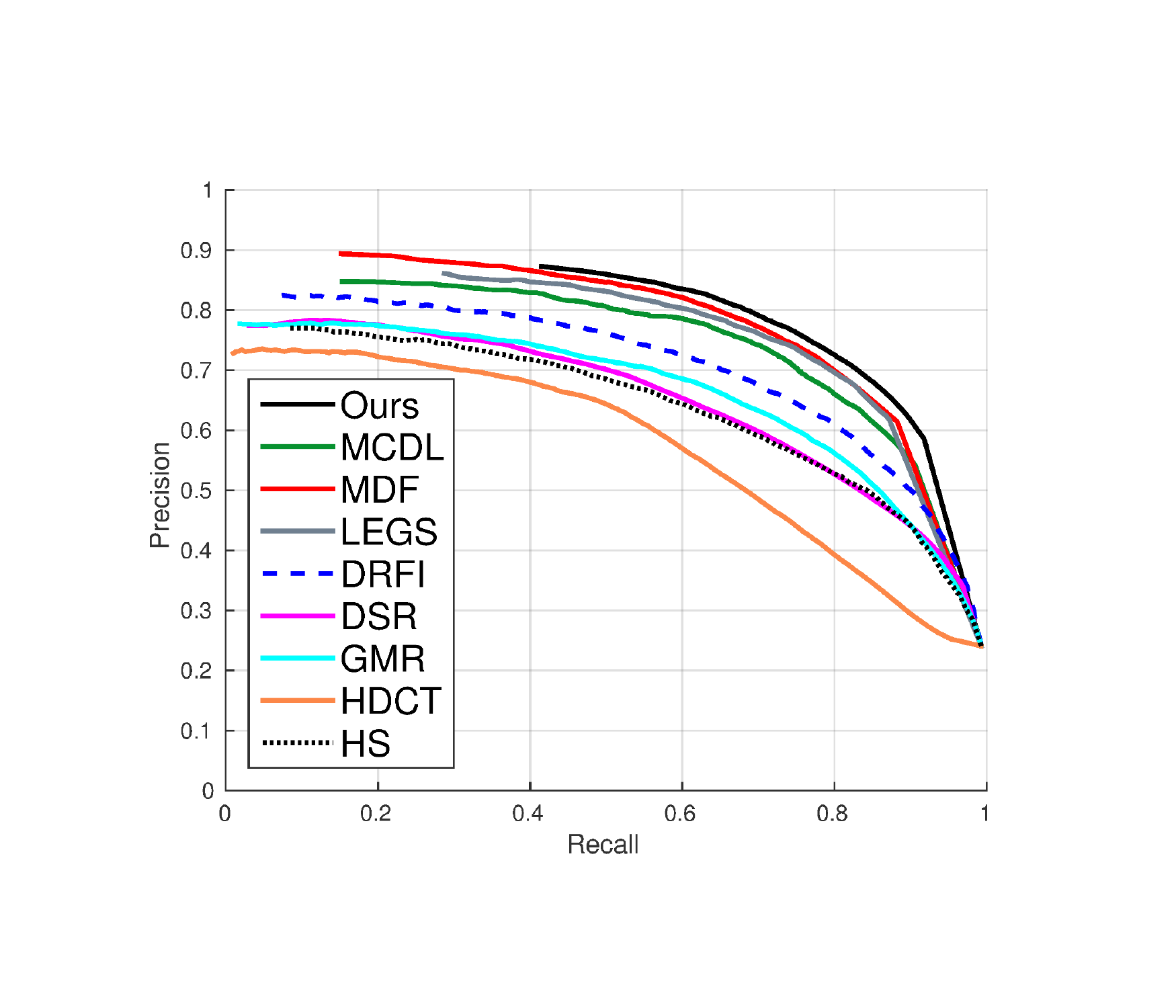}} &
\subfloat{\includegraphics[width=0.26\textwidth]{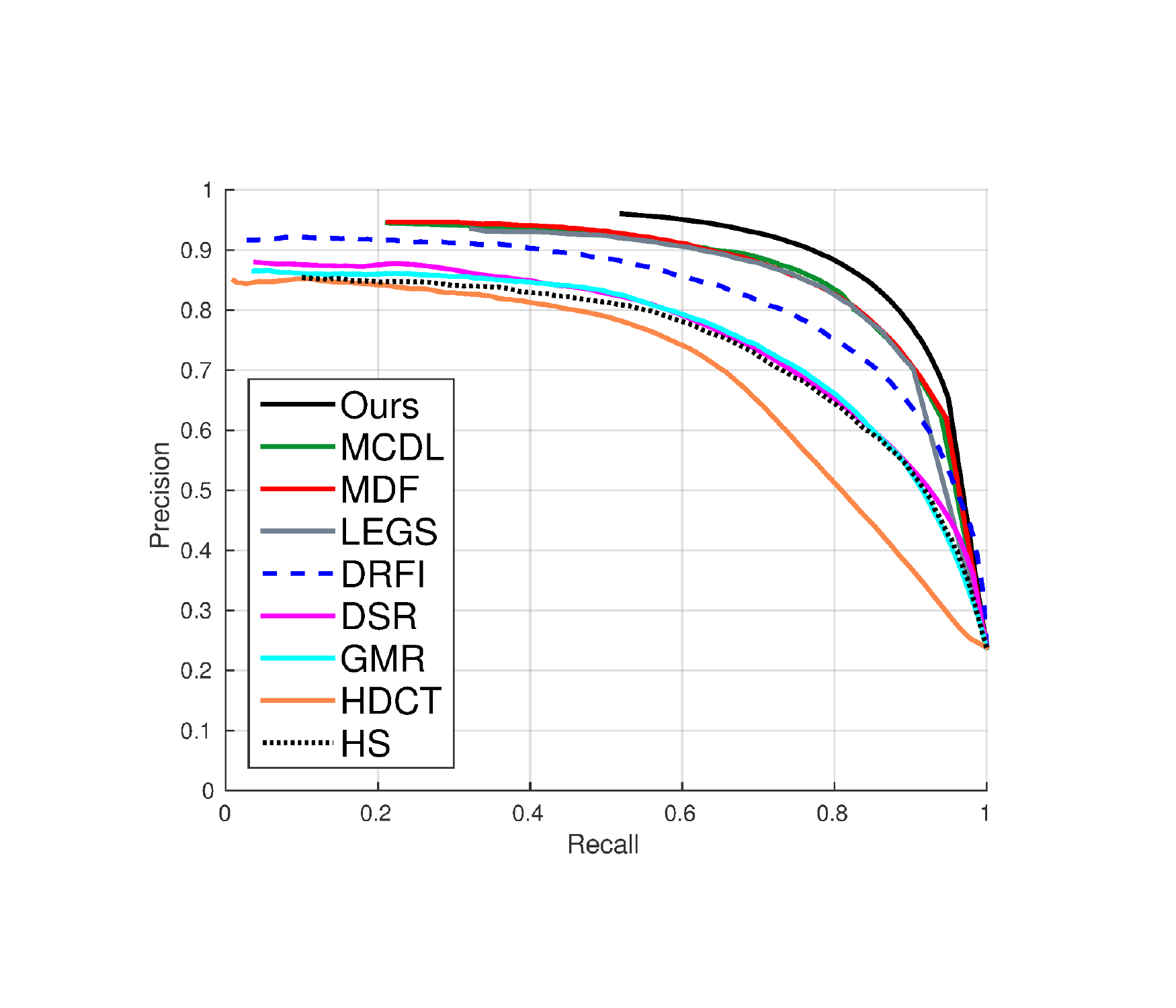}} &
\subfloat{\includegraphics[width=0.26\textwidth]{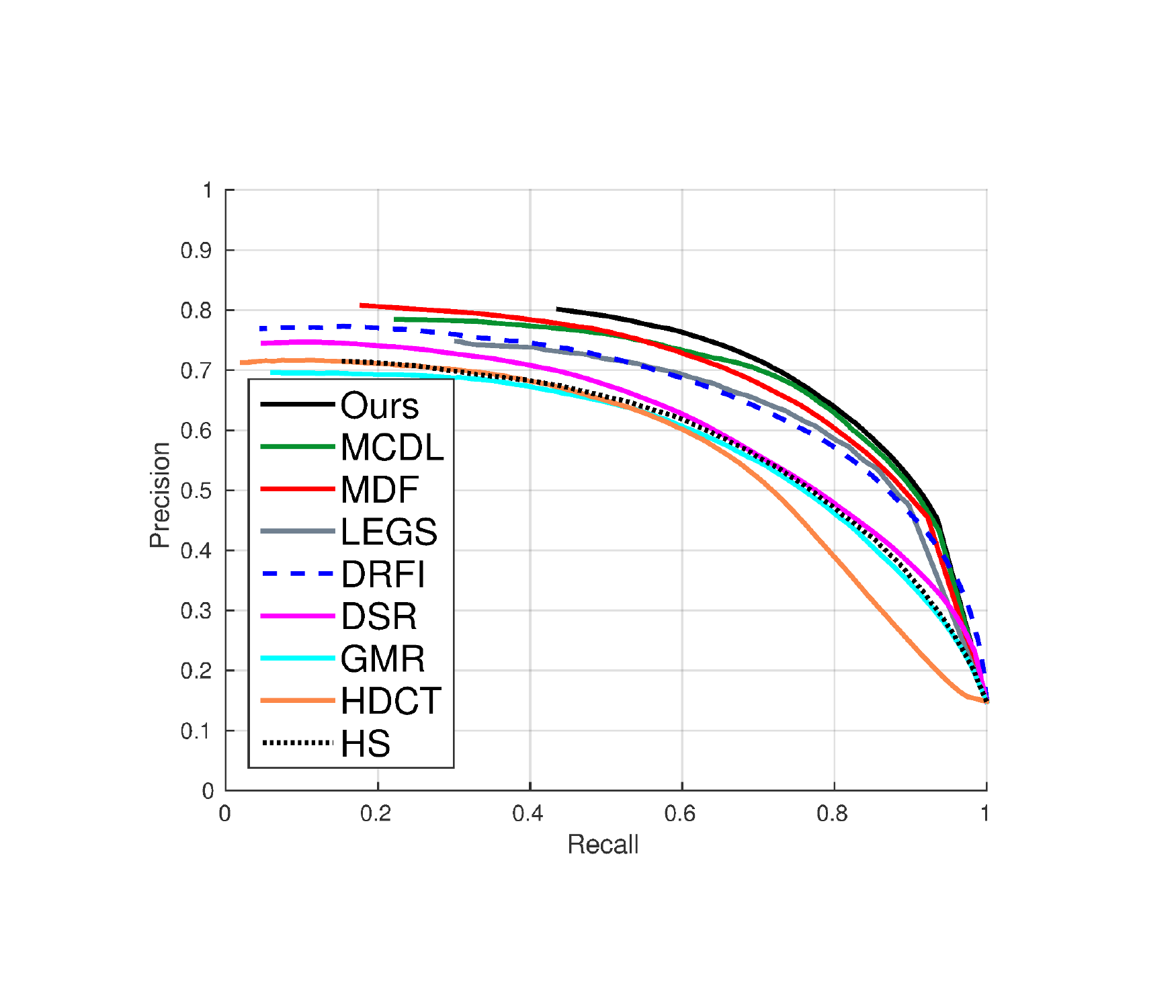}} &
\subfloat{\includegraphics[width=0.26\textwidth]{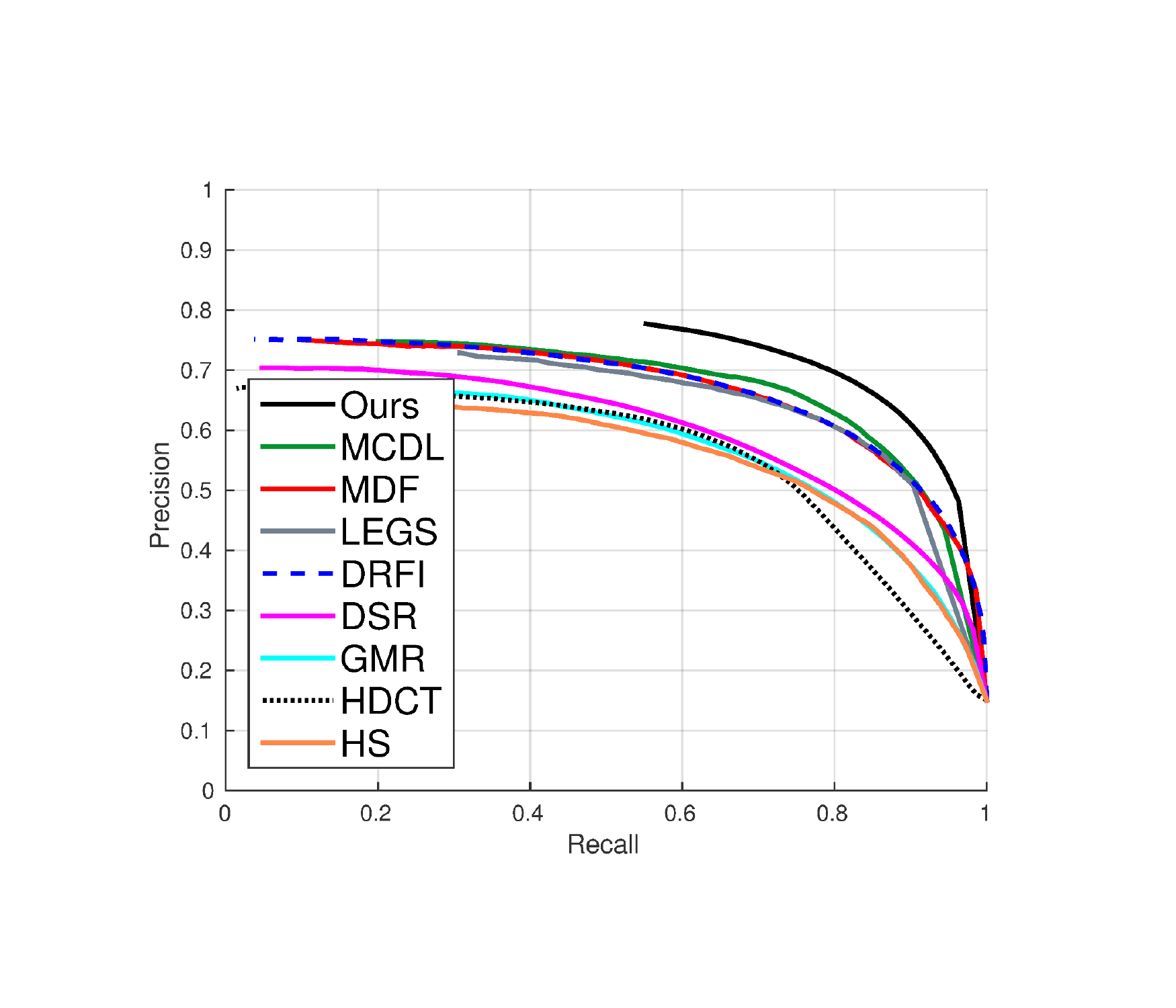}} \vspace{-0.6in}\\
\subfloat{\includegraphics[width=0.26\textwidth]{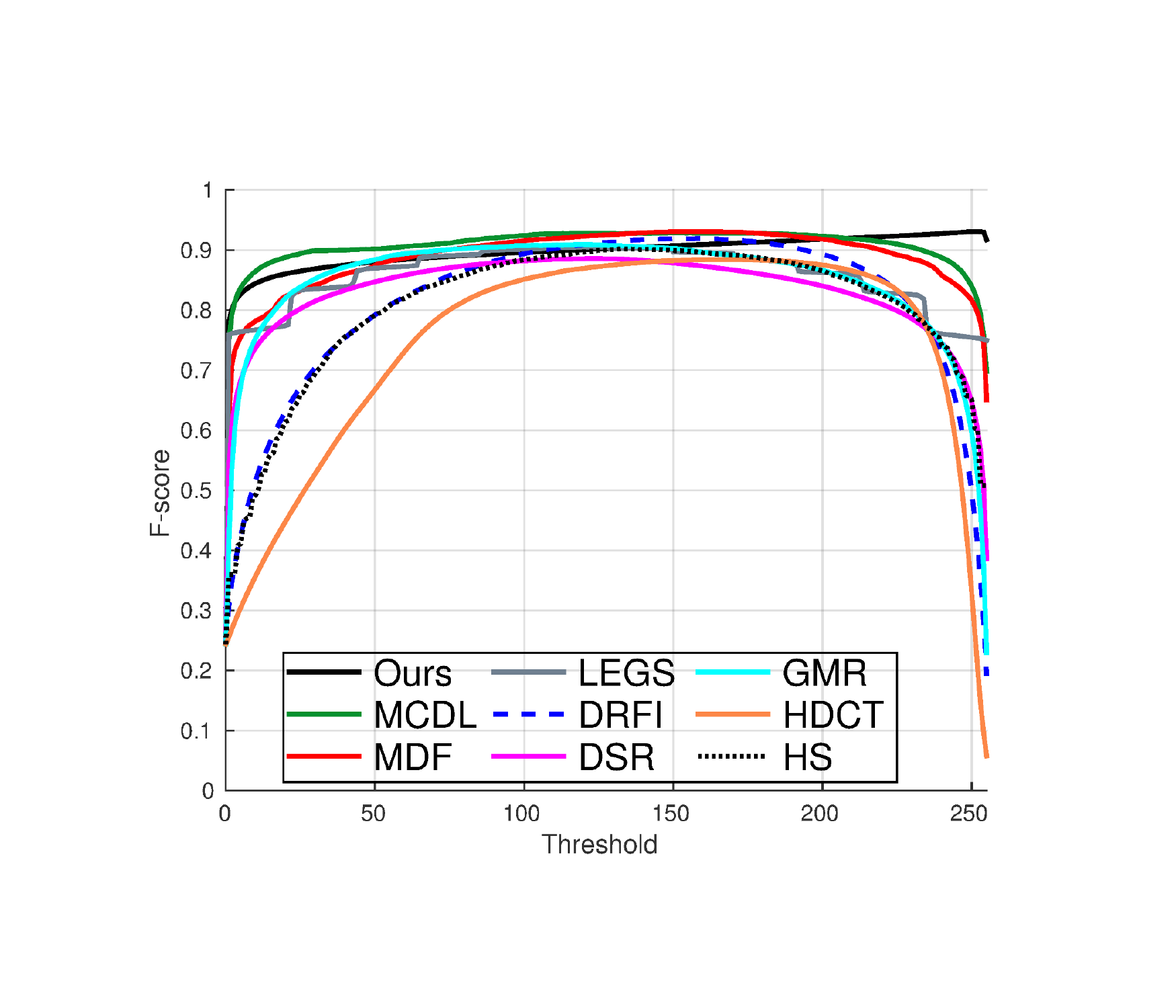}} &
\subfloat{\includegraphics[width=0.26\textwidth]{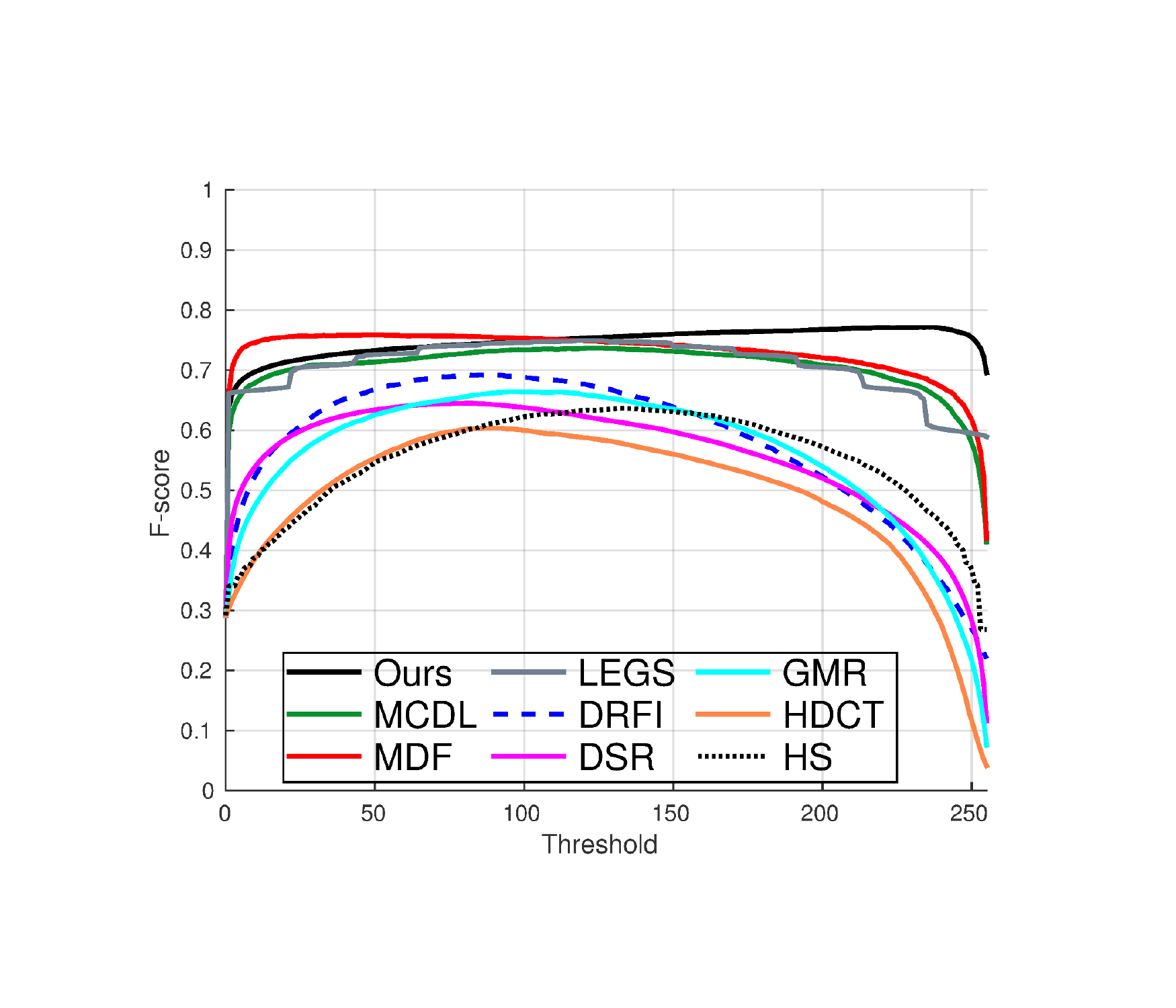}} &
\subfloat{\includegraphics[width=0.26\textwidth]{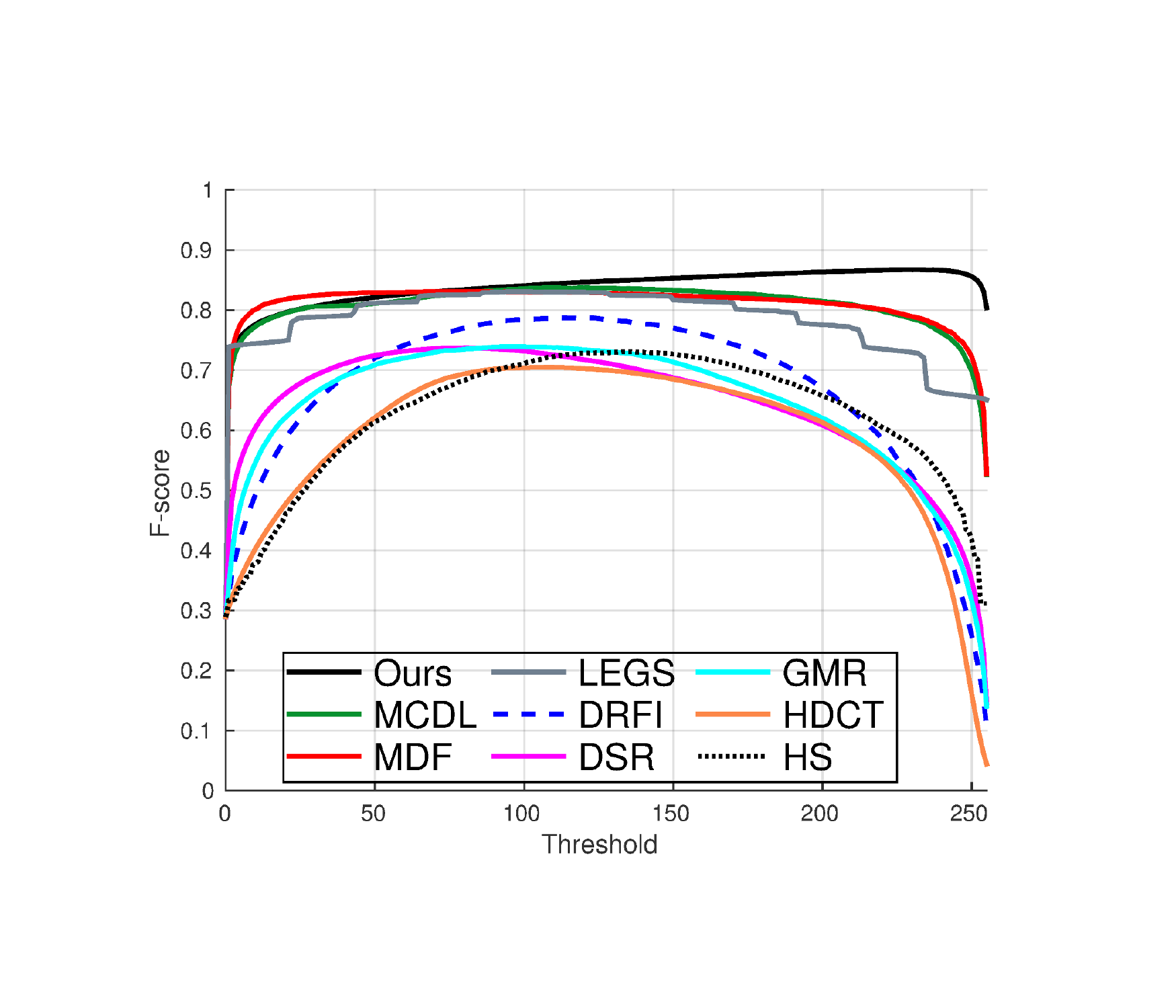}} &
\subfloat{\includegraphics[width=0.26\textwidth]{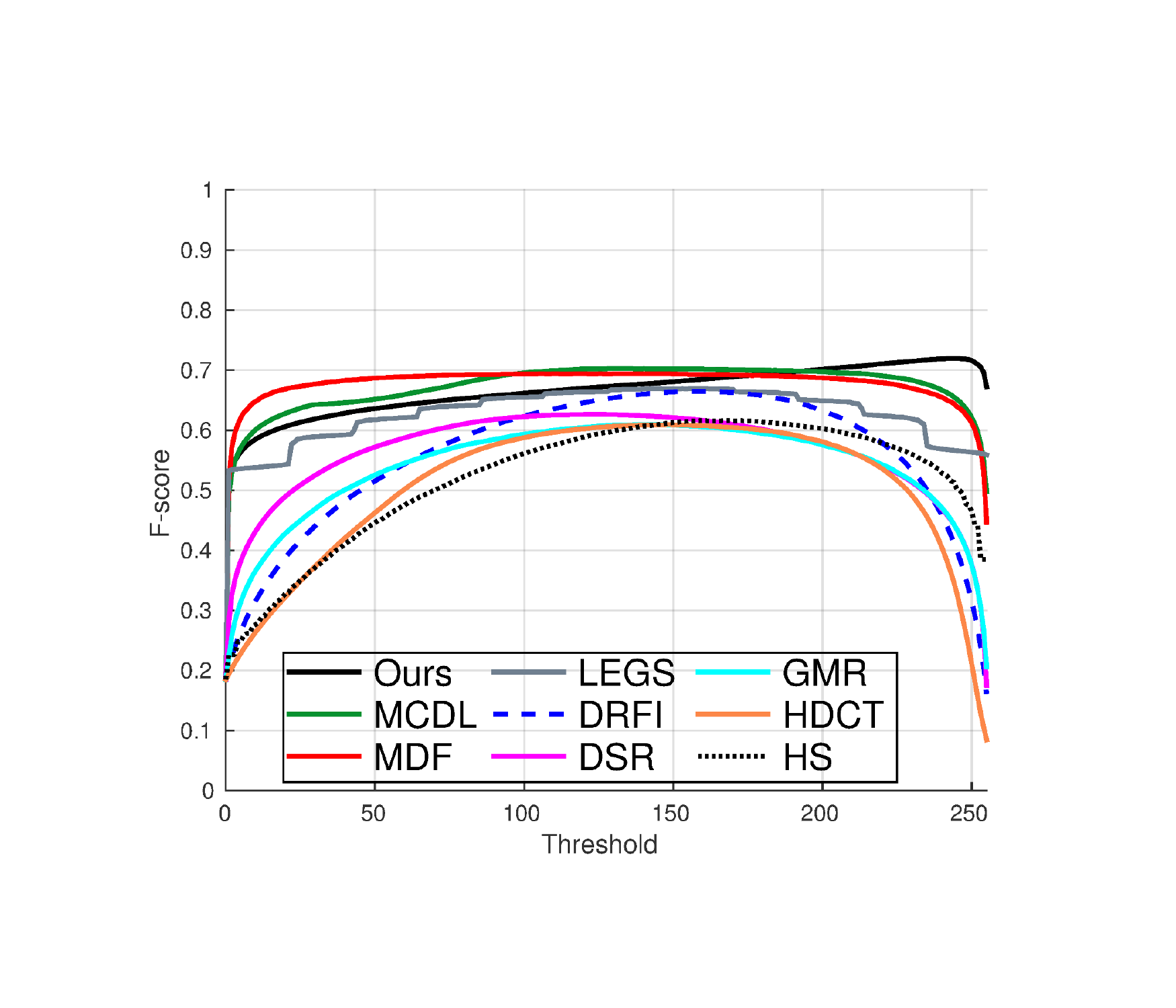}} &
\subfloat{\includegraphics[width=0.26\textwidth]{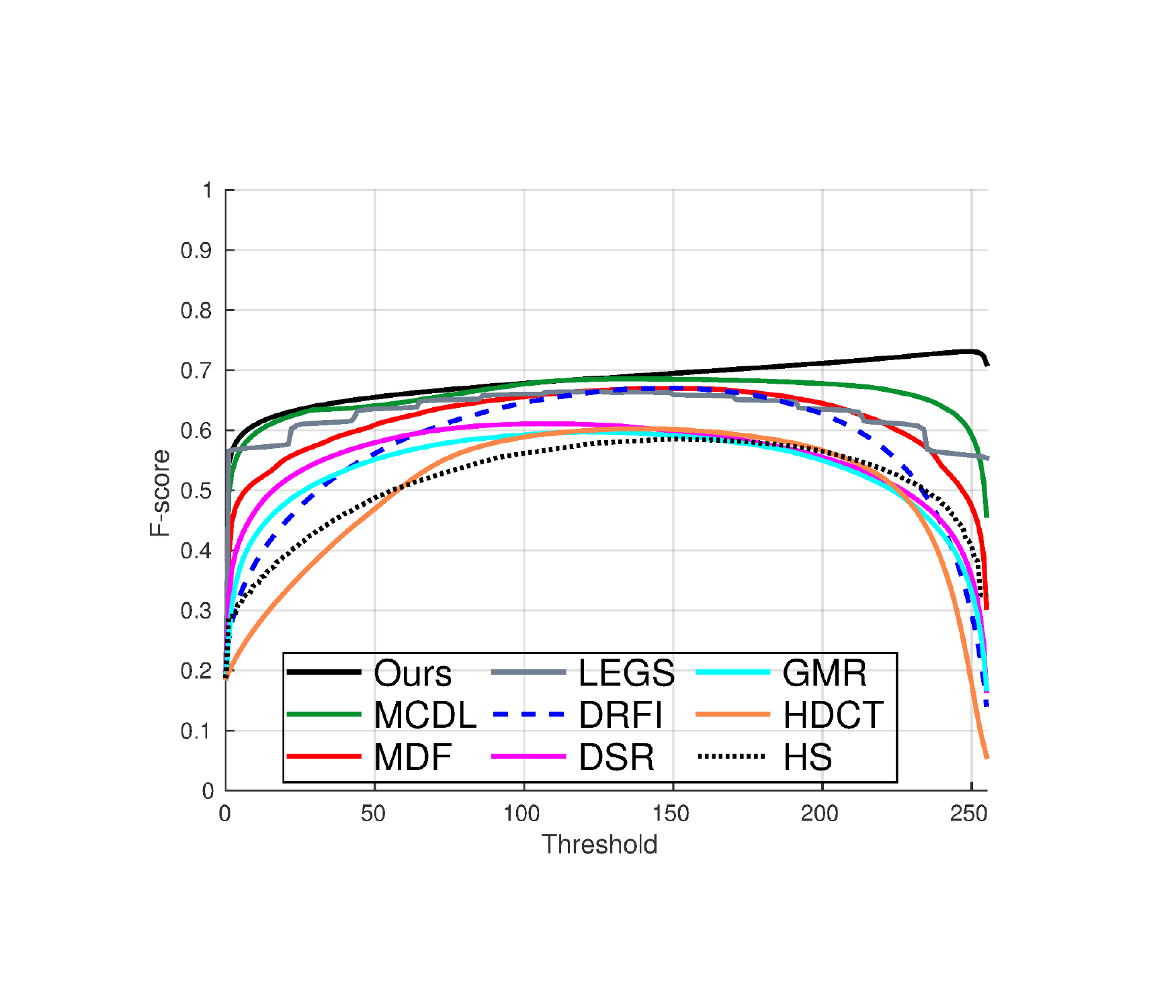}} \vspace{-0.5in}\\
\subfloat{\includegraphics[width=0.26\textwidth]{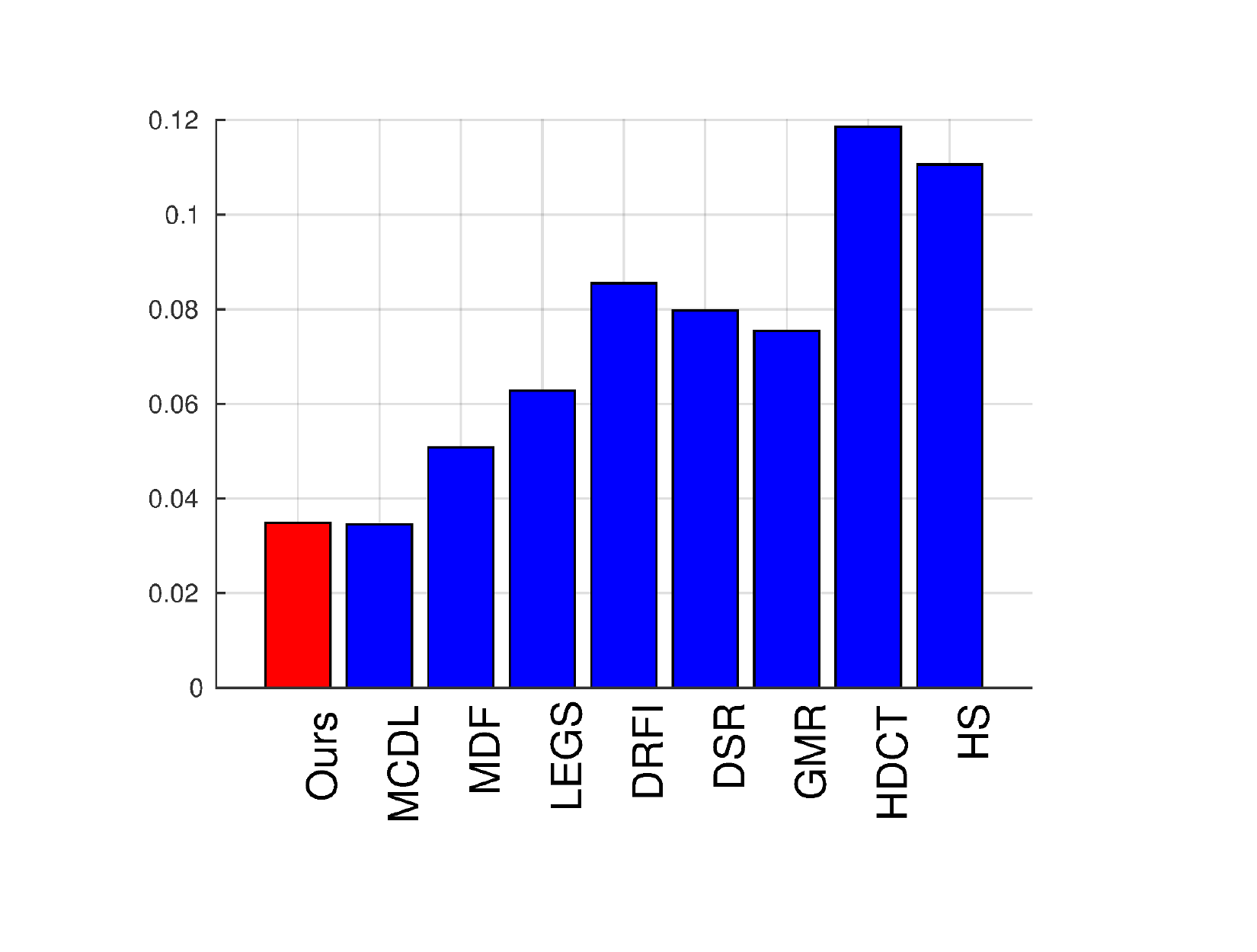}} &
\subfloat{\includegraphics[width=0.26\textwidth]{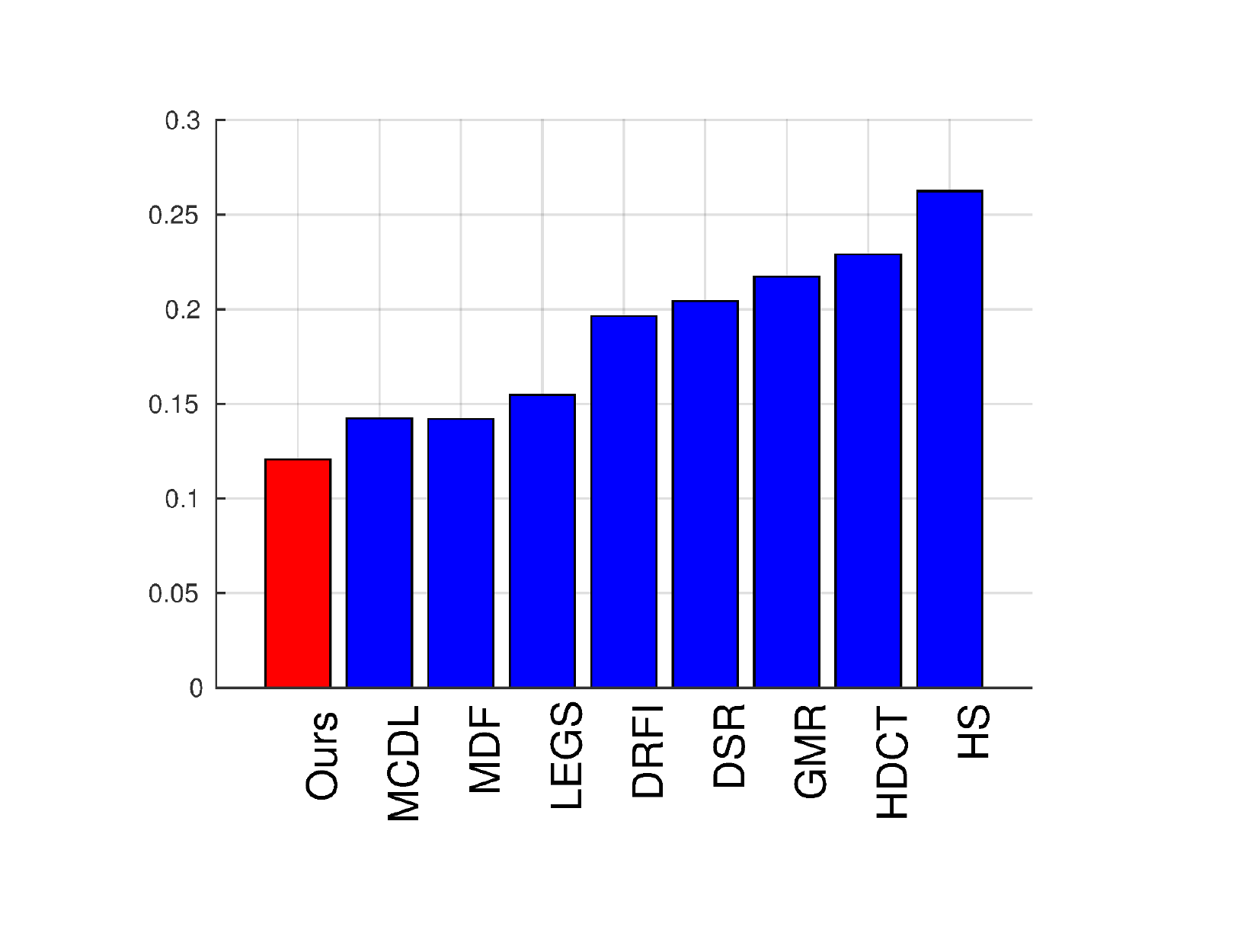}} &
\subfloat{\includegraphics[width=0.26\textwidth]{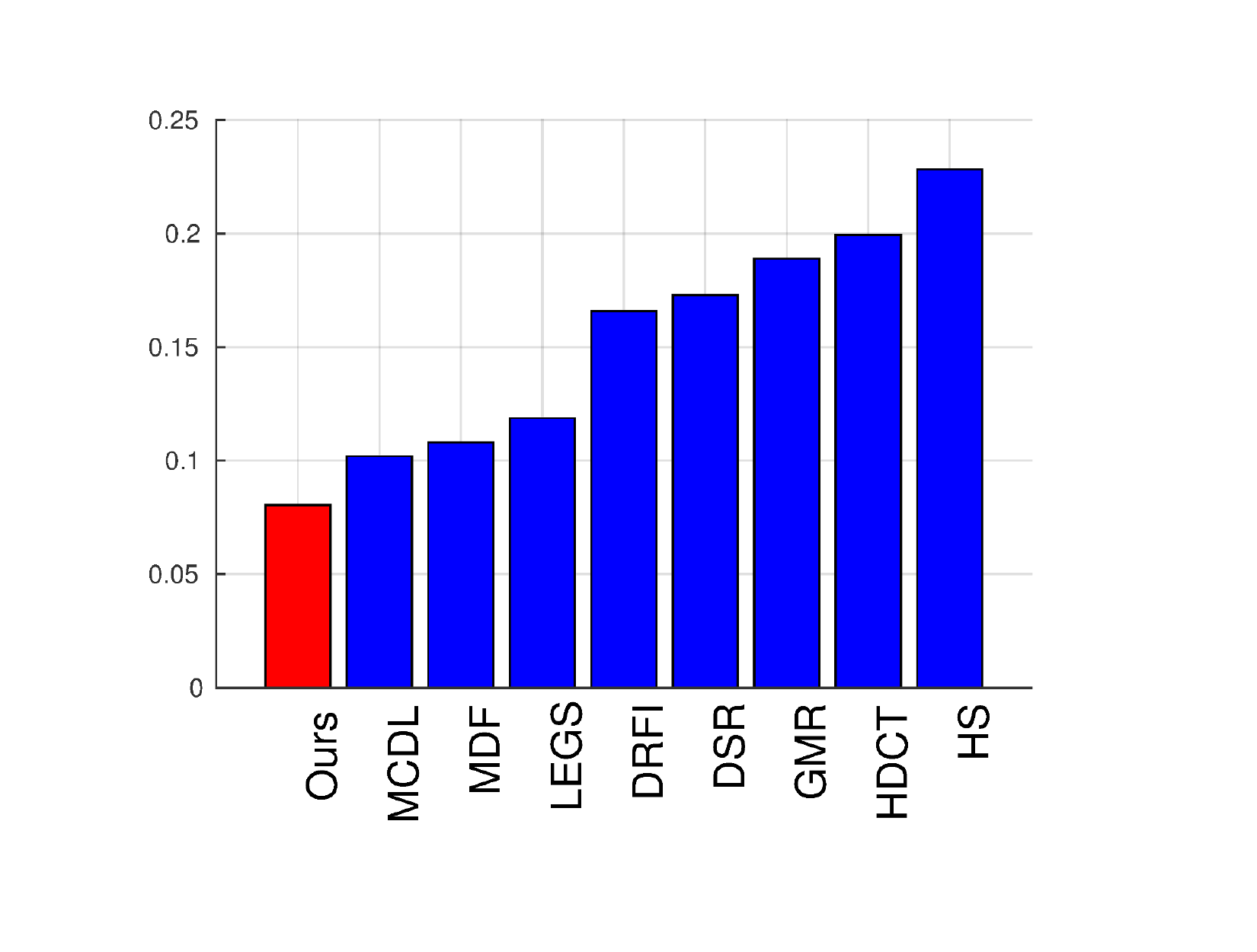}} &
\subfloat{\includegraphics[width=0.26\textwidth]{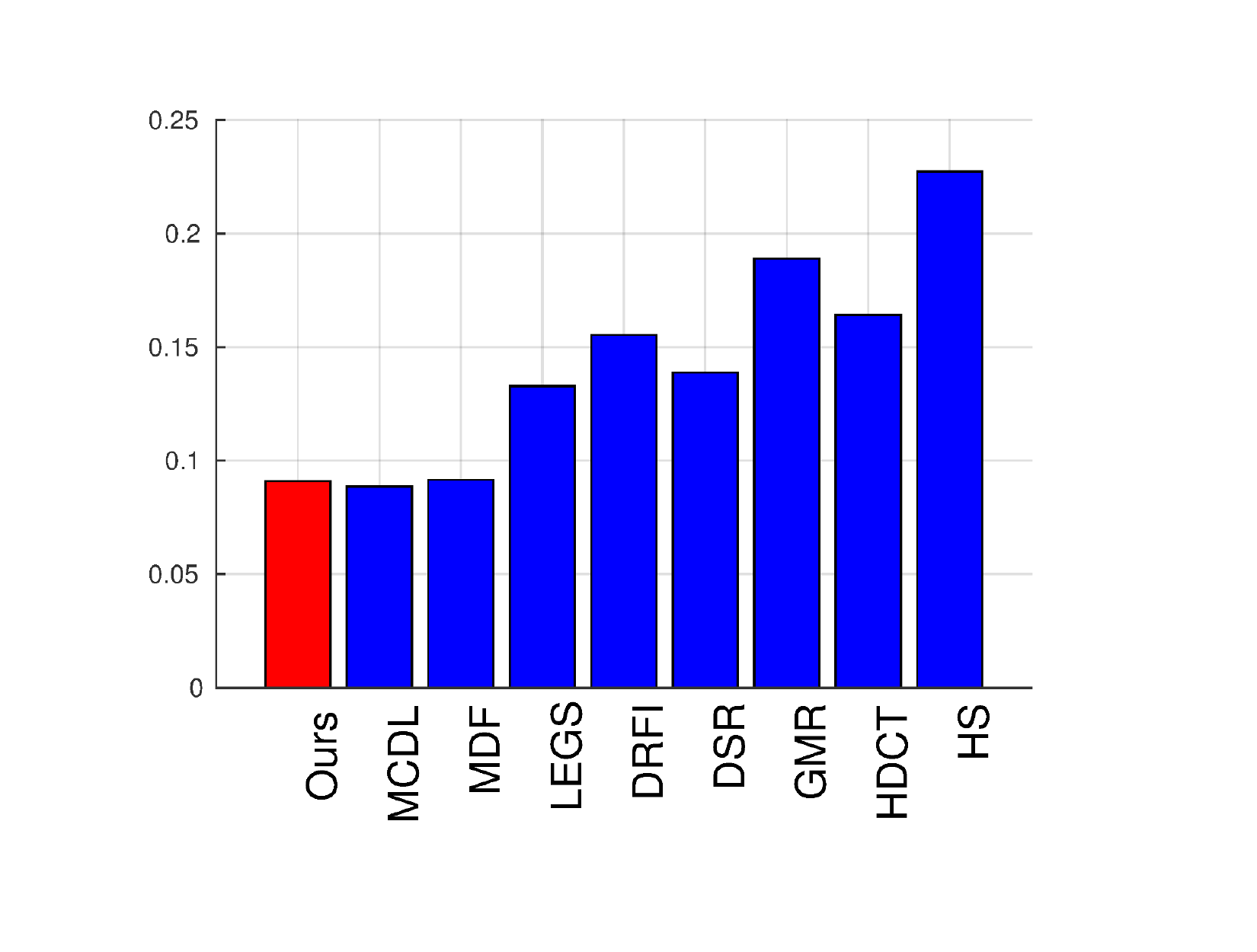}} &
\subfloat{\includegraphics[width=0.26\textwidth]{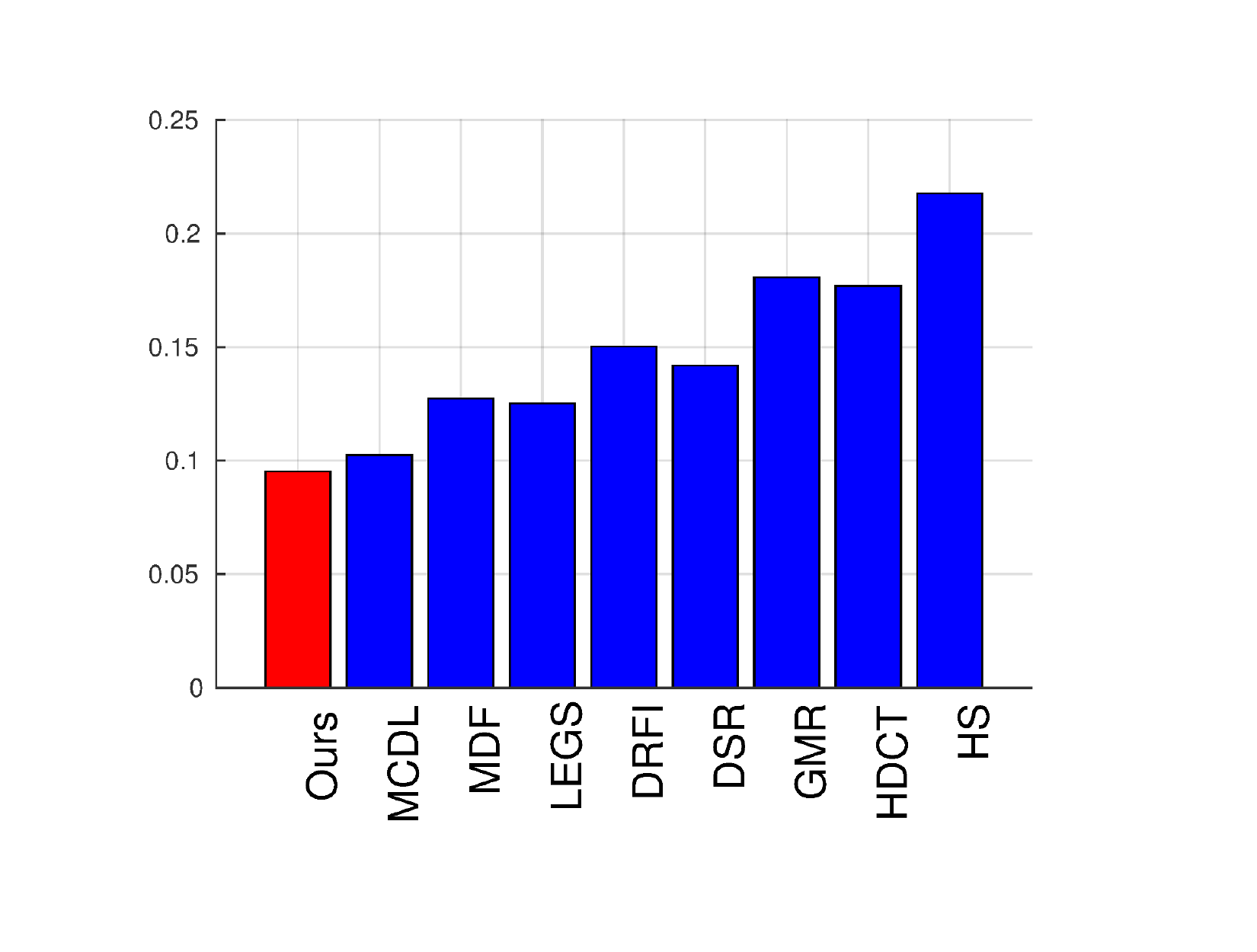}} \\
\setlength\tabcolsep{1pt}
(a) ASD &(b) PASCAL-S &(c) ECSSD &(d) DUT-OMRON &(e) THUR15K
\end{tabular} \vspace{-0.1in}
\caption{From top to bottom, Precision-Recall (PR) graph, F-measure score with different thresholds and Mean Absolute Error (MAE) of various algorithms on five popular datasets.}\vspace{-0.15in}\label{fig:compare}
\end{figure*}

\begin{table}
\scriptsize
\begin{center}
\begin{tabular}{|c|c|c|c|c|c|}
\hline
 & ASD & PASCAL-S & ECSSD & DUT-OMRON & THUR15K\\
\hline
Ours & 0.924 & \textbf{0.771} & \textbf{0.867} & \textbf{0.719} & \textbf{0.731}\\
\hline
MCDL & 0.928 & 0.737 & 0.837 & 0.703 & 0.686\\
\hline
MDF & \textbf{0.931} & 0.759 & 0.831 & 0.694 & 0.670\\
\hline
LEGS & 0.905 & 0.749 & 0.831 & 0.669 & 0.664\\
\hline
DRFI & 0.919 & 0.692 & 0.787 & 0.665 & 0.670\\
\hline
DSR & 0.886 & 0.645 & 0.737 & 0.626 & 0.611\\
\hline
GMR & 0.909 & 0.664 & 0.740 & 0.610 & 0.597\\
\hline
HDCT & 0.884 & 0.604 & 0.705 & 0.609 & 0.602\\
\hline
HS & 0.902 & 0.637 & 0.731 & 0.616 & 0.585\\
\hline
\end{tabular}
\end{center}
\vspace{-0.15in}
\caption{The F-measure scores of salient region detection algorithms on five popular datasets. The best score is marked in bold.}\vspace{-0.1in}
\label{tab:result_table_fscore}
\end{table}
\begin{table}
\scriptsize
\begin{center}
\begin{tabular}{|c|c|c|c|c|c|}
\hline
 & ASD & PASCAL-S & ECSSD & DUT-OMRON & THUR15K\\
\hline
Ours & \textbf{0.035} & \textbf{0.121} & \textbf{0.080} & 0.091 & \textbf{0.095}\\
\hline
MCDL & \textbf{0.035} & 0.142 & 0.102 & \textbf{0.089} & 0.102\\
\hline
MDF & 0.051 & 0.142 & 0.108 & 0.092 & 0.127\\
\hline
LEGS & 0.063 & 0.155 & 0.119 & 0.133 & 0.125\\
\hline
DRFI & 0.085 & 0.196 & 0.166 & 0.155 & 0.150\\
\hline
DSR & 0.080 & 0.205 & 0.173 & 0.139 & 0.142\\
\hline
GMR & 0.075 & 0.217 & 0.189 & 0.189 & 0.181\\
\hline
HDCT & 0.119 & 0.229 & 0.199 & 0.164 & 0.177\\
\hline
HS & 0.111 & 0.262 & 0.228 & 0.227 & 0.218\\
\hline
\end{tabular}
\end{center}
\vspace{-0.15in}
\caption{The Mean Absolute Error(MAE) of salient region detection algorithms on five popular datasets. The best score is marked in bold.}\vspace{-0.15in}
\label{tab:result_table_mae}
\end{table}

We evaluated the performance of our algorithm using various datasets. The \textbf{MSRA10K}~\cite{msra10k} is a dataset with 10,000 images which includes the \textbf{ASD} dataset~\cite{Achanta09cvpr}. Most images in this dataset contains single object. The \textbf{PASCAL-S}~\cite{pascals} is generated from the PASCAL VOC dataset~\cite{pascal} and contains 850 natural images. The \textbf{ECSSD}~\cite{hs} contains 1,000 images which have semantic meaning in their ground truth segmentation. It also contains images with complex structures. The \textbf{DUT-OMRON}~\cite{gmr} has 5,168 high quality images and the \textbf{THUR15K}~\cite{thur15k} contains 6,232 images of specific classes. 

We trained our model using 9,000 images from the MSRA10K dataset after excluding the same images in ASD dataset. We did not use validation set and trained the model until its training data loss converges. From each image, we use about 30 salient superpixels and 70 non-salient superpixels; around 0.9 million input data are generated. \rev{The layers of VGG16 model are fixed by setting the learning rate equal to zero. For other layers, we initialize the weights by the ``xavier'' (caffe parameter), and we set the base learning rate equal to 0.001.} We use stochastic gradient descent method with momentum 0.9 and decrease running rate 90\% when training loss does not decrease. Training our model takes 3 hours for 100,000 iterations with mini-batch size 128.

Our results were compared with MCDL~\cite{mcdl}, MDF~\cite{mdf}, LEGS~\cite{legs}, DRFI~\cite{drfi}, DSR~\cite{dsr}, GMR~\cite{gmr}, HDCT~\cite{hdct}, and HS~\cite{hs}, which are the state-of-the-art algorithms. DRFI, DSR, GMR, HDCT and HS use low level features and MCDL, MDF and LEGS utilize deep CNN for high level context. We obtained the result images from the project site of each algorithm or the benchmark evaluation~\cite{SalObjBenchmark}. The results which were not provided were generated from the authors' source codes published in the web. The comparisons on Precision-Recall(PR) graph and Mean Absolute Error(MAE) graph are presented in \Figref{compare}. Maximum F-measure scores and MAE values are also described in \Tabref{result_table_fscore} and \Tabref{result_table_mae}. We used the evaluation codes used in the benchmark paper~\cite{SalObjBenchmark}. The PR graph and f-measure score tend to be more informative than ROC curve because salient pixels are usually less than non-salient~\cite{SalObjBenchmark}. Following the criteria by Achanta et. al.~\cite{Achanta09cvpr}, we moved the threshold from 0 to 255 to generate binary masks($M$). Using the ground truth($G$), the precision and recall is calculated as follows:

\begin{equation}
Precision = \frac{\left | M\bigcap G \right |}{\left | M \right |},\quad
Recall = \frac{\left | M\bigcap G \right |}{\left | G \right |}
\end{equation}

We also reported the F-Measure score which is a balanced measurement between precision and recall as follows:

\begin{equation}
F_\beta  = \frac{(1+\beta^2)Precision \times Recall}{\beta^2 \times Precision + Recall}
\end{equation}
where $\beta^2$ is typically set to 0.3. We visualized f-measure score for the different thresholds and reported the maximum f-measure score which well describes the overall detection performance~\cite{SalObjBenchmark}. In our algorithm, making binary masks using the high threshold around 240 generated good f-measure score.

The overlapping-based evaluations give higher score to methods which assign high saliency score to salient pixel correctly. However, the evaluation on non-salient regions can be unfair especially for the methods which successfully detect non-salient regions, but missed the detection of salient regions~\cite{SalObjBenchmark}. Therefore, we also calculated the mean absolute error(MAE) for fair comparisons as suggested by~\cite{SalObjBenchmark}. The MAE evaluates
the detection accuracy as follow:
\begin{equation}
MAE = \frac{1}{W\times H}\sum_{x=1}^{W}\sum_{y=1}^{H}\left | S(x,y) - G(x,y)  \right |
\end{equation}
where $W$ and $H$ are width and height of an image, $S$ is the estimated saliency map and $G$ is the ground truth binary mask.

\begin{figure*}
\centering
\setlength\tabcolsep{1pt}
\begin{tabular}{ccccccccccc}
\subfloat{\includegraphics[width=0.084\textwidth]{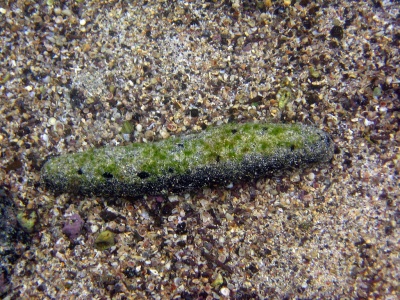}} &
\subfloat{\includegraphics[width=0.084\textwidth]{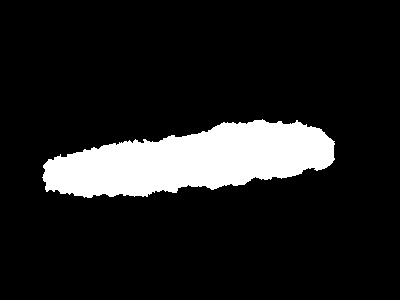}} &
\subfloat{\includegraphics[width=0.084\textwidth]{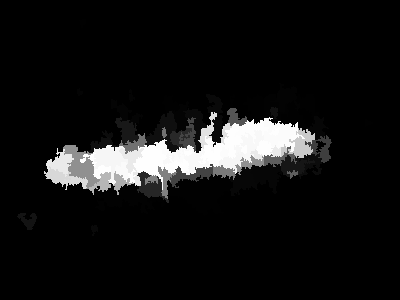}} &
\subfloat{\includegraphics[width=0.084\textwidth]{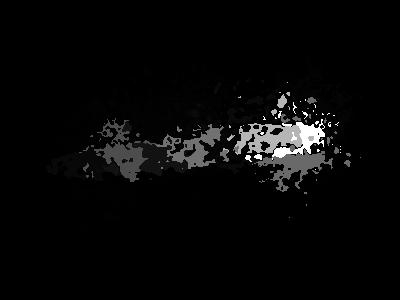}} &
\subfloat{\includegraphics[width=0.084\textwidth]{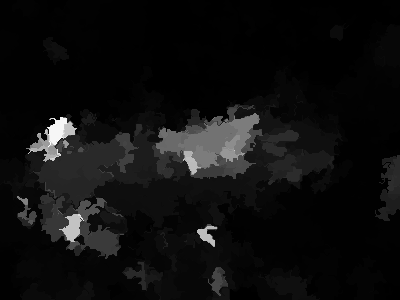}} &
\subfloat{\includegraphics[width=0.084\textwidth]{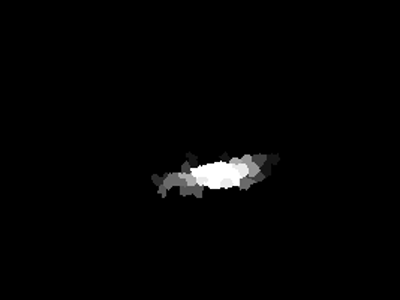}} &
\subfloat{\includegraphics[width=0.084\textwidth]{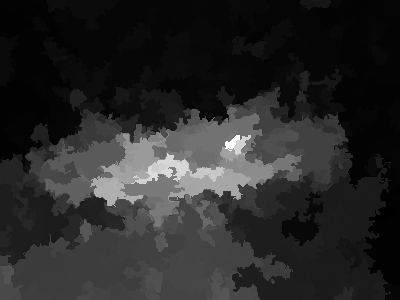}} &
\subfloat{\includegraphics[width=0.084\textwidth]{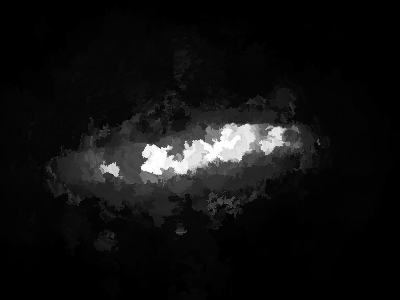}} &
\subfloat{\includegraphics[width=0.084\textwidth]{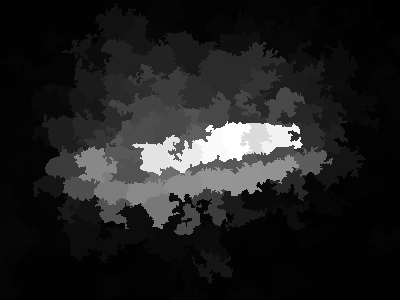}} &
\subfloat{\includegraphics[width=0.084\textwidth]{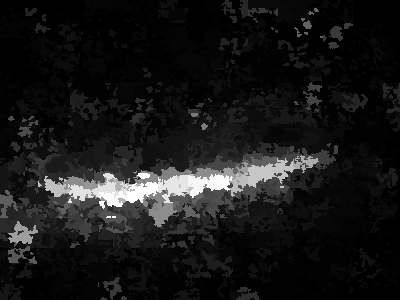}} &
\subfloat{\includegraphics[width=0.084\textwidth]{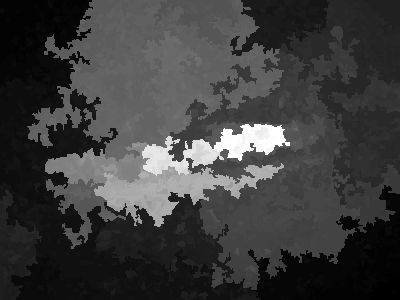}} \vspace{-0.15in}\\

\subfloat{\includegraphics[width=0.084\textwidth]{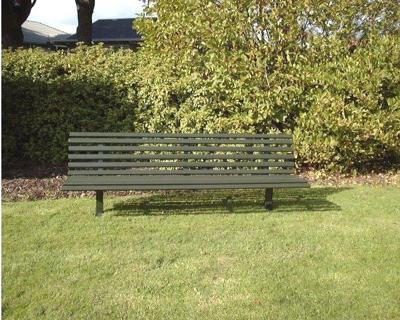}} &
\subfloat{\includegraphics[width=0.084\textwidth]{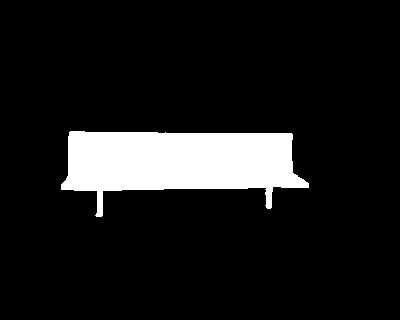}} &
\subfloat{\includegraphics[width=0.084\textwidth]{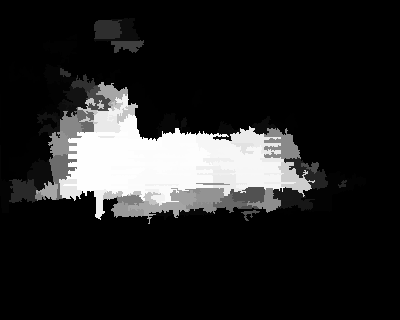}} &
\subfloat{\includegraphics[width=0.084\textwidth]{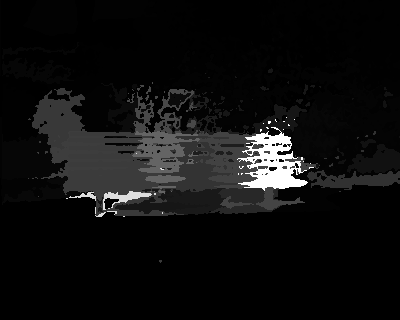}} &
\subfloat{\includegraphics[width=0.084\textwidth]{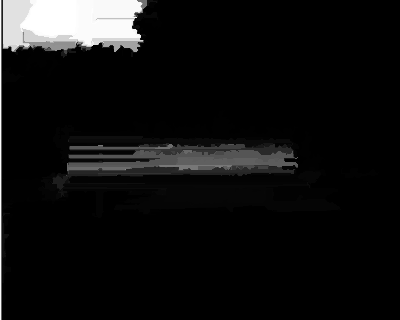}} &
\subfloat{\includegraphics[width=0.084\textwidth]{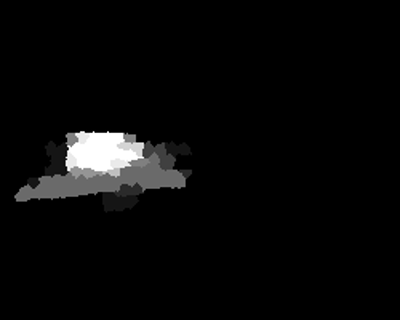}} &
\subfloat{\includegraphics[width=0.084\textwidth]{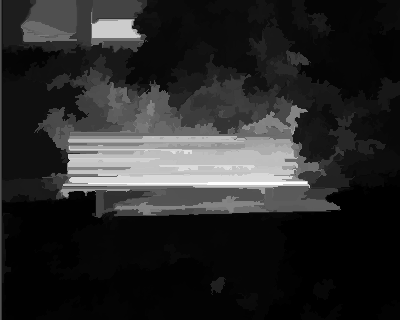}} &
\subfloat{\includegraphics[width=0.084\textwidth]{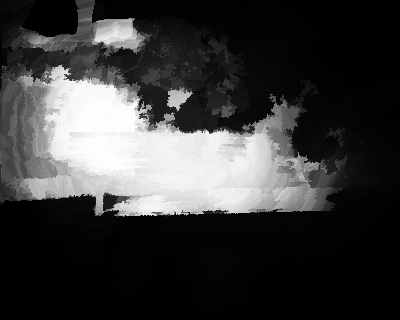}} &
\subfloat{\includegraphics[width=0.084\textwidth]{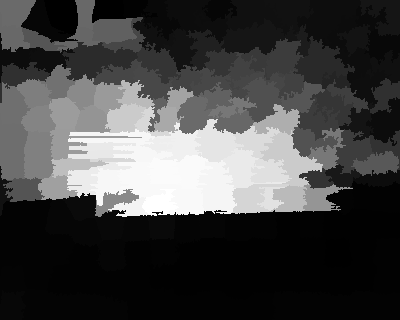}} &
\subfloat{\includegraphics[width=0.084\textwidth]{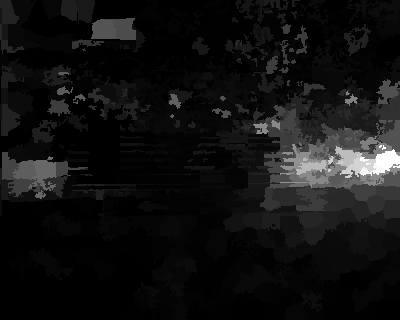}} &
\subfloat{\includegraphics[width=0.084\textwidth]{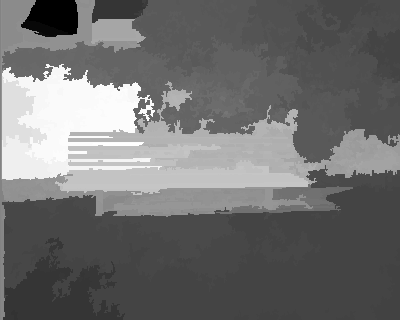}} \vspace{-0.15in}\\

\subfloat{\includegraphics[width=0.084\textwidth]{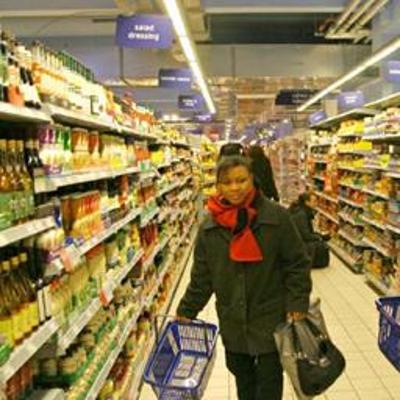}} &
\subfloat{\includegraphics[width=0.084\textwidth]{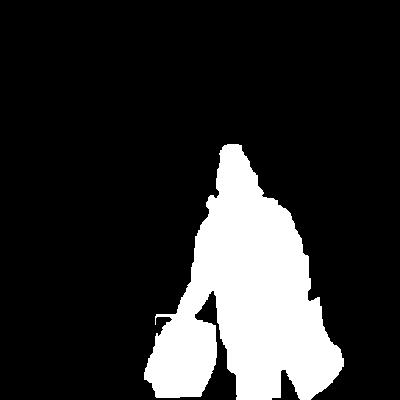}} &
\subfloat{\includegraphics[width=0.084\textwidth]{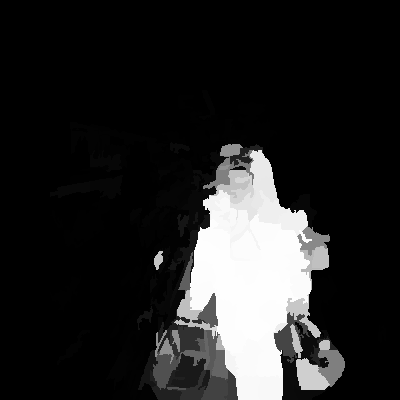}} &
\subfloat{\includegraphics[width=0.084\textwidth]{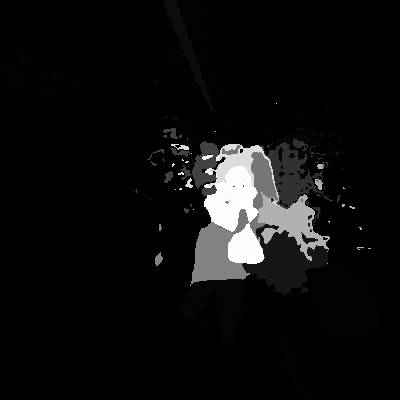}} &
\subfloat{\includegraphics[width=0.084\textwidth]{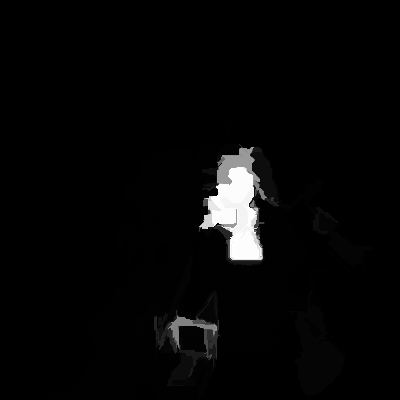}} &
\subfloat{\includegraphics[width=0.084\textwidth]{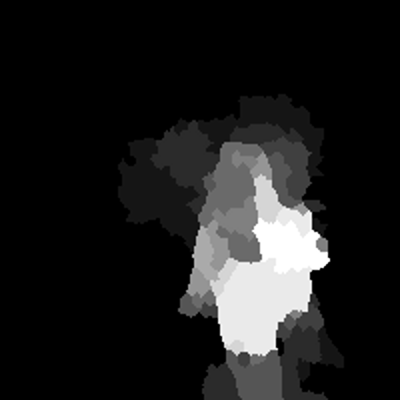}} &
\subfloat{\includegraphics[width=0.084\textwidth]{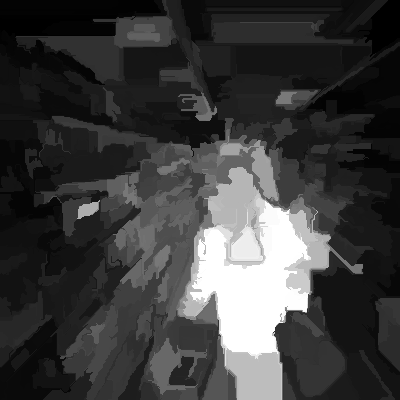}} &
\subfloat{\includegraphics[width=0.084\textwidth]{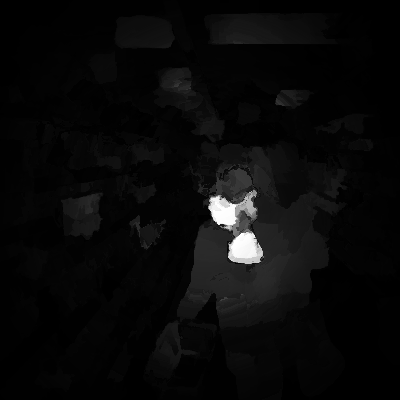}} &
\subfloat{\includegraphics[width=0.084\textwidth]{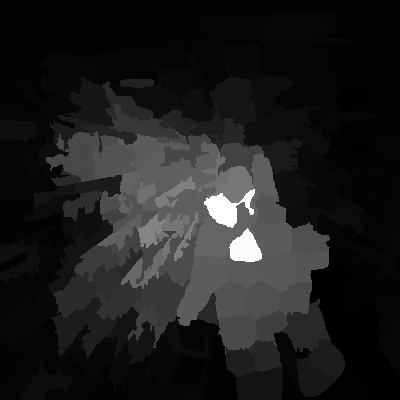}} &
\subfloat{\includegraphics[width=0.084\textwidth]{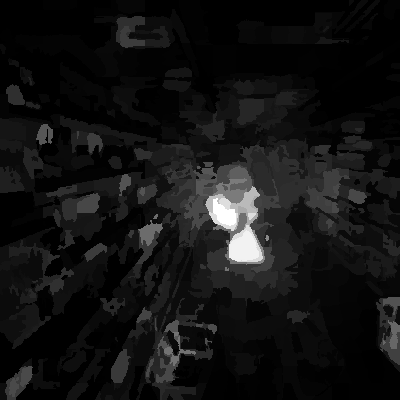}} &
\subfloat{\includegraphics[width=0.084\textwidth]{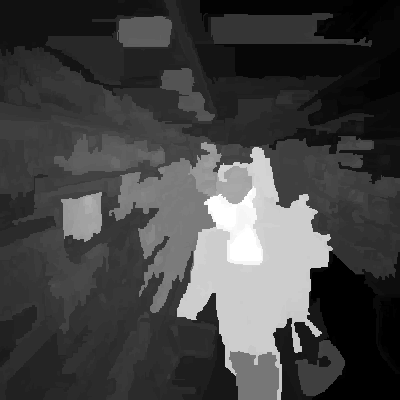}}  \vspace{-0.15in}\\

\subfloat{\includegraphics[width=0.084\textwidth]{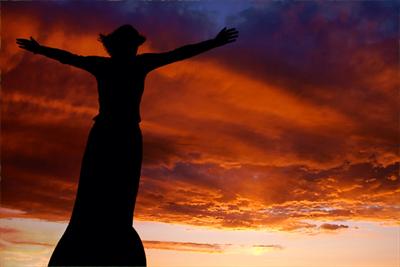}} &
\subfloat{\includegraphics[width=0.084\textwidth]{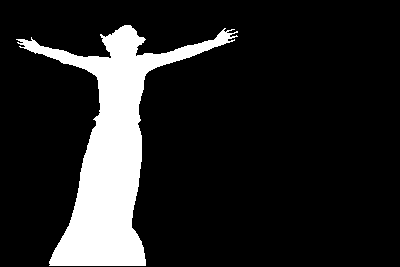}} &
\subfloat{\includegraphics[width=0.084\textwidth]{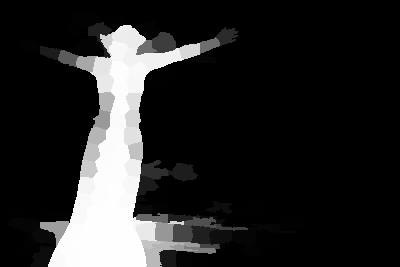}} &
\subfloat{\includegraphics[width=0.084\textwidth]{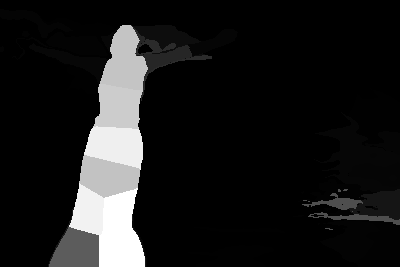}} &
\subfloat{\includegraphics[width=0.084\textwidth]{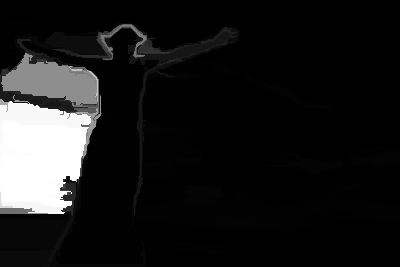}} &
\subfloat{\includegraphics[width=0.084\textwidth]{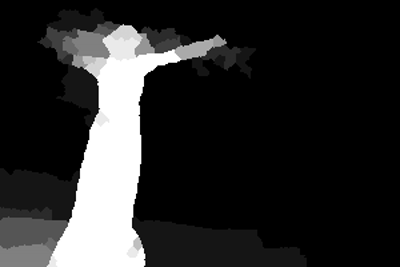}} &
\subfloat{\includegraphics[width=0.084\textwidth]{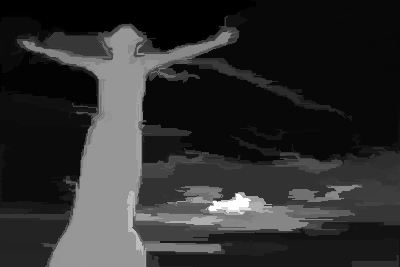}} &
\subfloat{\includegraphics[width=0.084\textwidth]{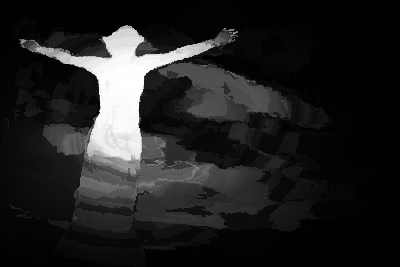}} &
\subfloat{\includegraphics[width=0.084\textwidth]{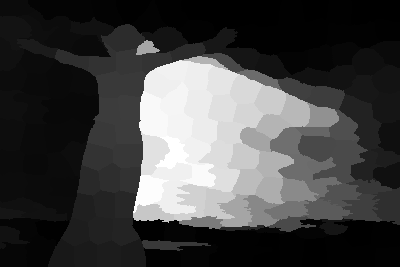}} &
\subfloat{\includegraphics[width=0.084\textwidth]{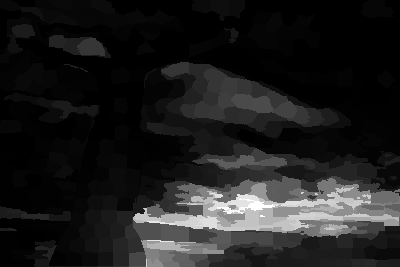}} &
\subfloat{\includegraphics[width=0.084\textwidth]{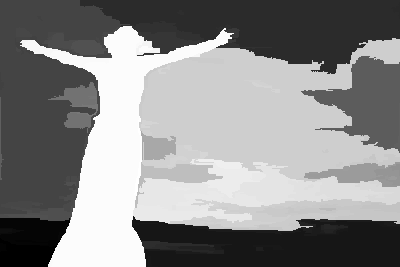}} \vspace{-0.15in}\\

\subfloat{\includegraphics[width=0.084\textwidth]{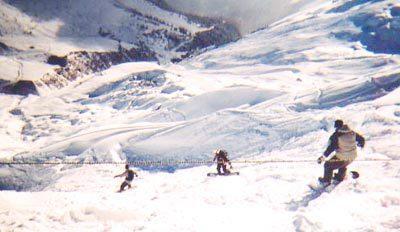}} &
\subfloat{\includegraphics[width=0.084\textwidth]{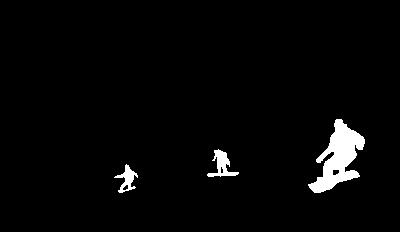}} &
\subfloat{\includegraphics[width=0.084\textwidth]{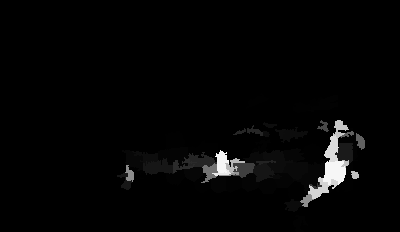}} &
\subfloat{\includegraphics[width=0.084\textwidth]{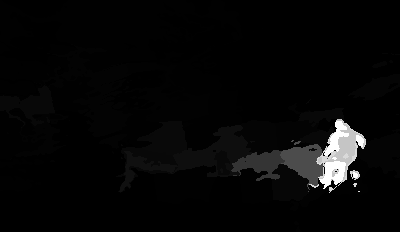}} &
\subfloat{\includegraphics[width=0.084\textwidth]{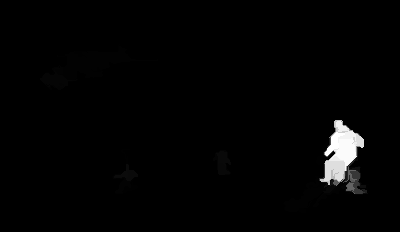}} &
\subfloat{\includegraphics[width=0.084\textwidth]{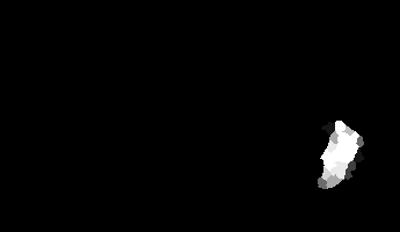}} &
\subfloat{\includegraphics[width=0.084\textwidth]{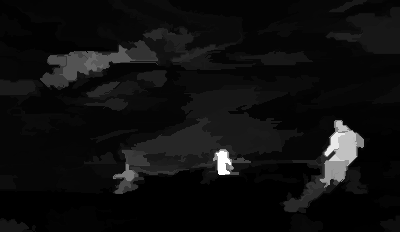}} &
\subfloat{\includegraphics[width=0.084\textwidth]{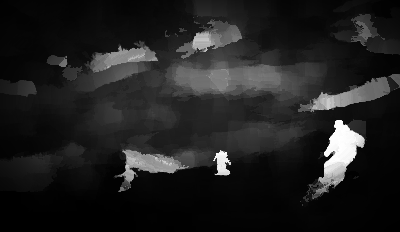}} &
\subfloat{\includegraphics[width=0.084\textwidth]{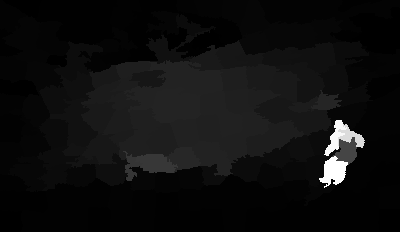}} &
\subfloat{\includegraphics[width=0.084\textwidth]{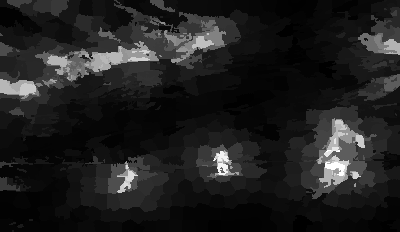}} &
\subfloat{\includegraphics[width=0.084\textwidth]{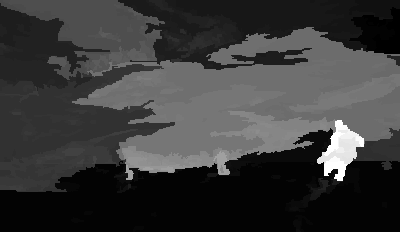}}  \vspace{-0.15in}\\

\subfloat{\includegraphics[width=0.084\textwidth]{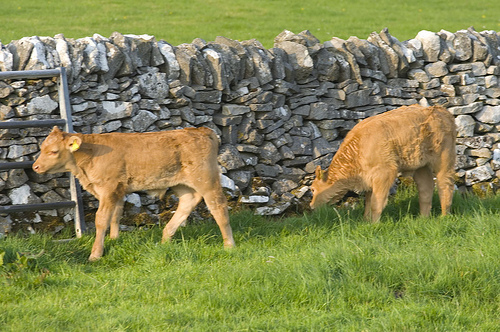}} &
\subfloat{\includegraphics[width=0.084\textwidth]{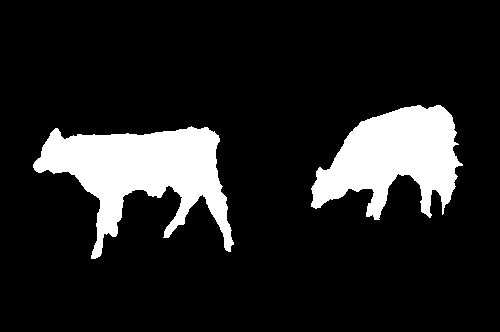}} &
\subfloat{\includegraphics[width=0.084\textwidth]{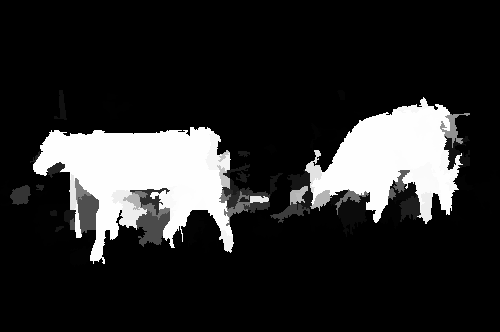}} &
\subfloat{\includegraphics[width=0.084\textwidth]{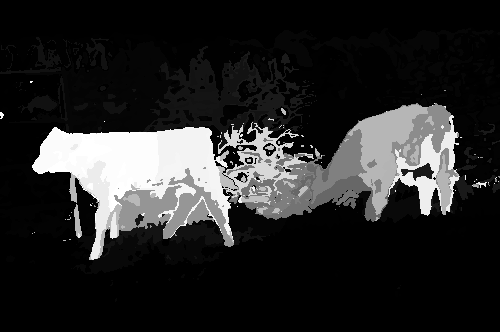}} &
\subfloat{\includegraphics[width=0.084\textwidth]{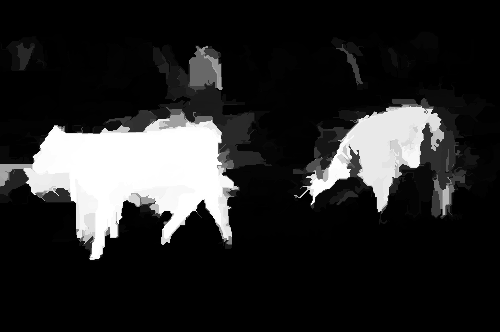}} &
\subfloat{\includegraphics[width=0.084\textwidth]{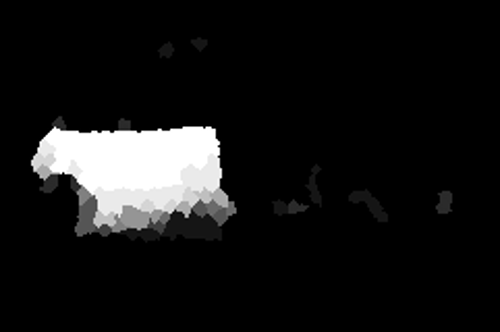}} &
\subfloat{\includegraphics[width=0.084\textwidth]{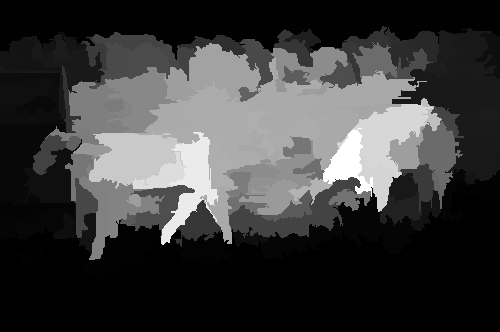}} &
\subfloat{\includegraphics[width=0.084\textwidth]{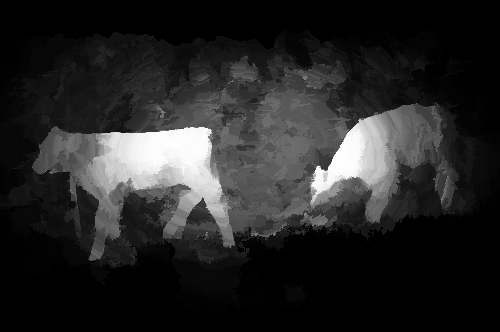}} &
\subfloat{\includegraphics[width=0.084\textwidth]{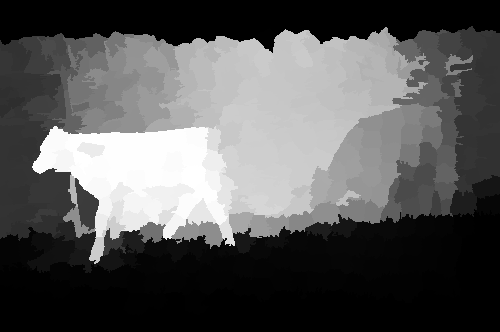}} &
\subfloat{\includegraphics[width=0.084\textwidth]{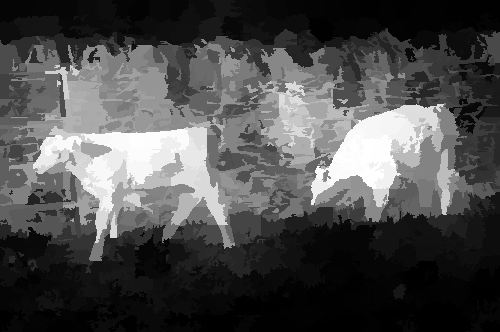}} &
\subfloat{\includegraphics[width=0.084\textwidth]{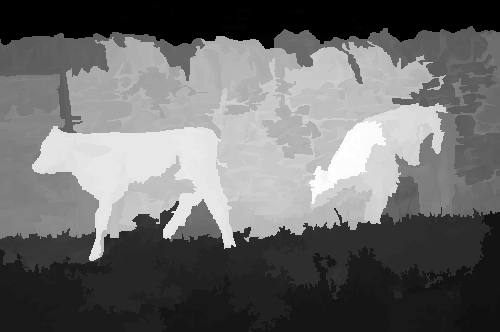}}  \vspace{-0.15in}\\

\subfloat{\includegraphics[width=0.084\textwidth]{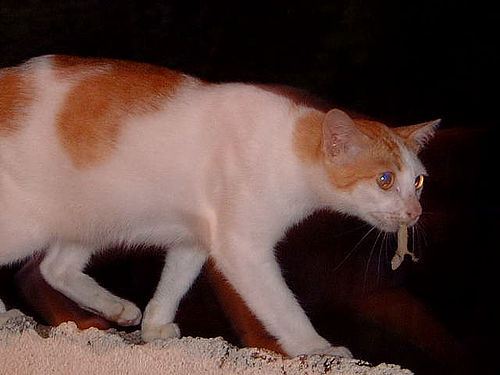}} &
\subfloat{\includegraphics[width=0.084\textwidth]{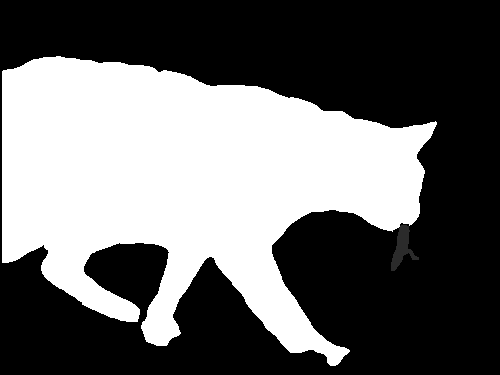}} &
\subfloat{\includegraphics[width=0.084\textwidth]{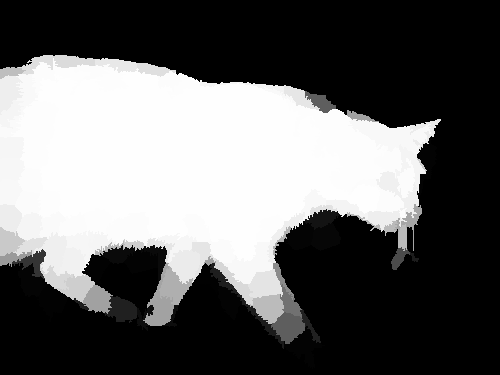}} &
\subfloat{\includegraphics[width=0.084\textwidth]{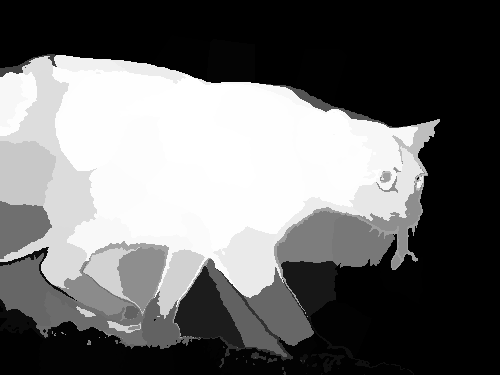}} &
\subfloat{\includegraphics[width=0.084\textwidth]{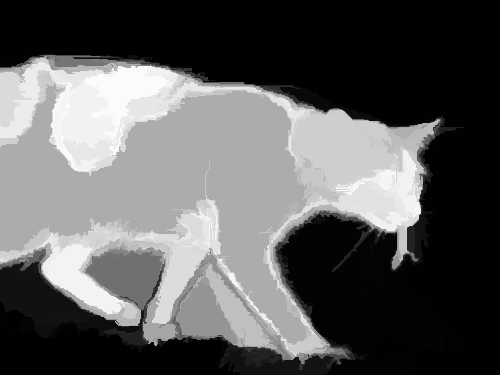}} &
\subfloat{\includegraphics[width=0.084\textwidth]{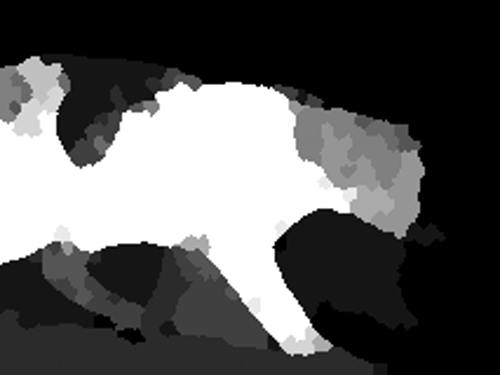}} &
\subfloat{\includegraphics[width=0.084\textwidth]{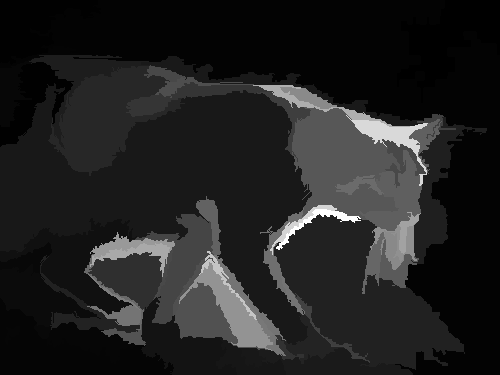}} &
\subfloat{\includegraphics[width=0.084\textwidth]{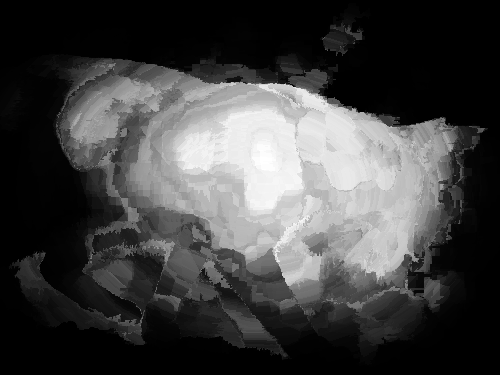}} &
\subfloat{\includegraphics[width=0.084\textwidth]{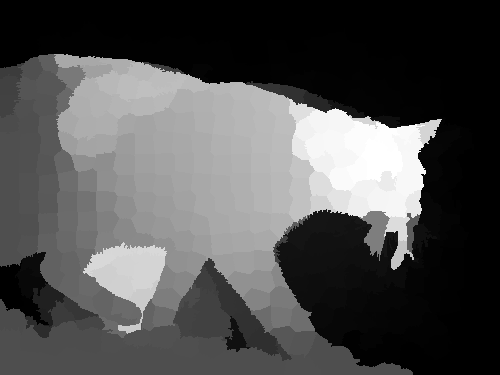}} &
\subfloat{\includegraphics[width=0.084\textwidth]{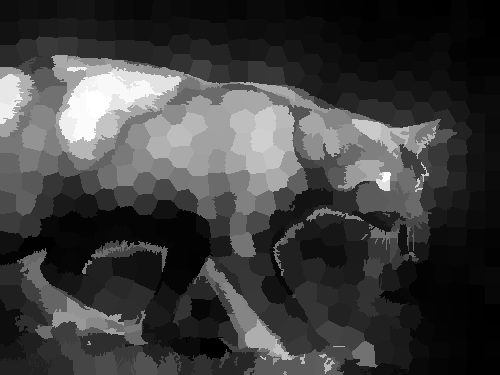}} &
\subfloat{\includegraphics[width=0.084\textwidth]{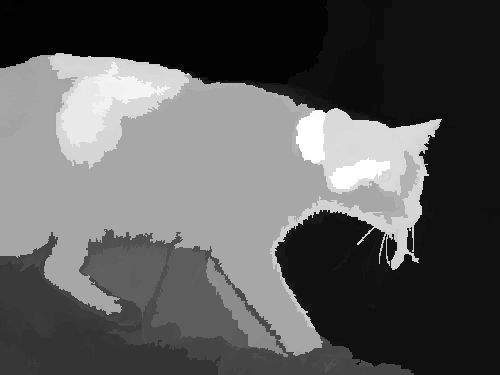}}\\

 (a) Inputs & (b) GT & (c) Ours & (d) MCDL & (e) MDF & (f) LEGS & (g) DRFI & (h) DSR & (i) GMR & (j) HDCT & (k) HS
\end{tabular} \vspace{-0.15in}
\caption{Visual comparisons of our results and the state-of-the-art methods on difficult scenes. (a) original image, (b) ground truth, (c) Ours (d) MCDL~\cite{mcdl} (e) MDF~\cite{mdf} (f) LEGS~\cite{legs} (g) DRFI~\cite{drfi}, (h) DSR~\cite{dsr}, (i) GMR~\cite{gmr}, (j) HDCT~\cite{hdct} (k) HS~\cite{hs}. From the top to the bottom, row 1-2 are the images with a low-contrast salient object, row 3-4 are with complicated background, row 5-6 are with multiple salient objects and row 7 is with a salient object touching the image boundaries.}\vspace{-0.15in}
\label{fig:visualize_compare2}
\end{figure*}

\begin{figure}
\centering
\setlength\tabcolsep{1pt}
\begin{tabular}{ccccc}
\subfloat{\includegraphics[width=0.19\linewidth]{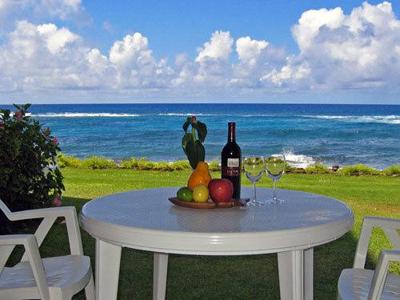}} &
\subfloat{\includegraphics[width=0.19\linewidth]{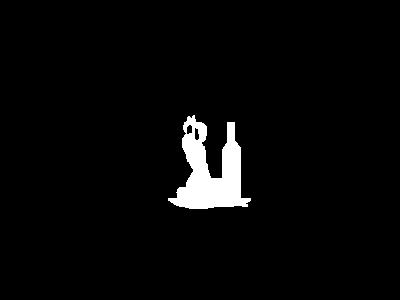}} &
\subfloat{\includegraphics[width=0.19\linewidth]{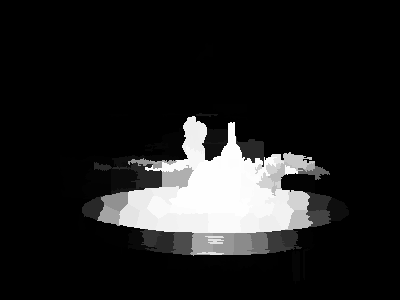}} &
\subfloat{\includegraphics[width=0.19\linewidth] {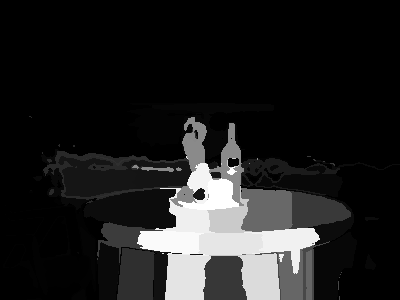}} &
\subfloat{\includegraphics[width=0.19\linewidth]{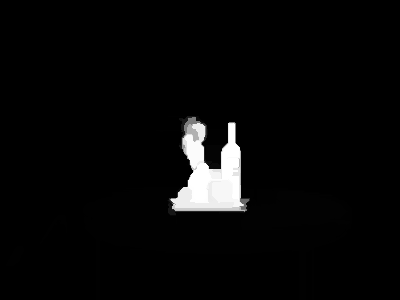}} \vspace{-0.15in}\\
\subfloat{\includegraphics[width=0.19\linewidth]{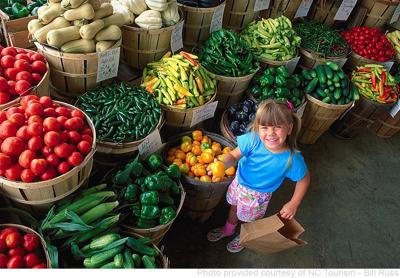}} &
\subfloat{\includegraphics[width=0.19\linewidth]{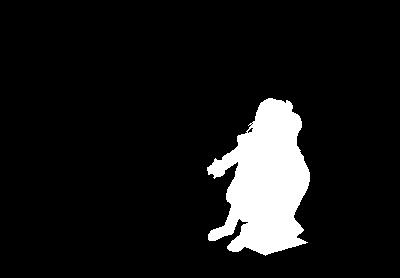}} &
\subfloat{\includegraphics[width=0.19\linewidth]{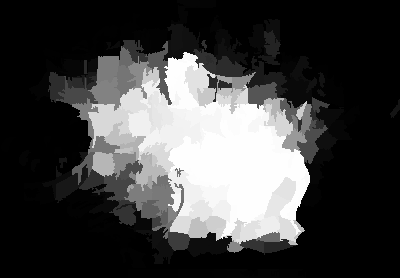}} &
\subfloat{\includegraphics[width=0.19\linewidth]{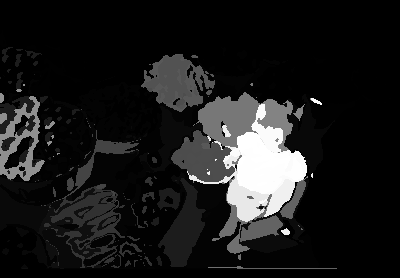}}&
\subfloat{\includegraphics[width=0.19\linewidth]{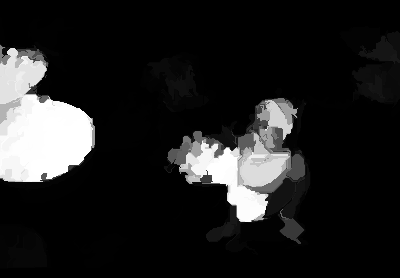}}\vspace{-0.15in}\\
\subfloat{\includegraphics[width=0.19\linewidth]{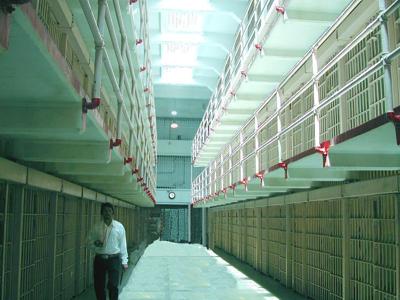}} &
\subfloat{\includegraphics[width=0.19\linewidth]{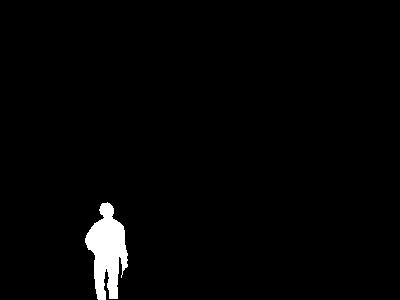}} &
\subfloat{\includegraphics[width=0.19\linewidth]{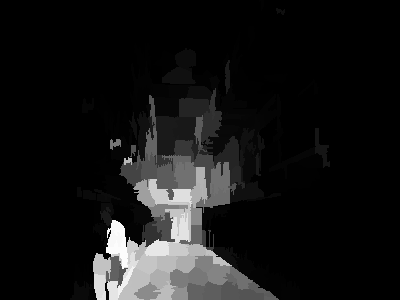}} &
\subfloat{\includegraphics[width=0.19\linewidth]{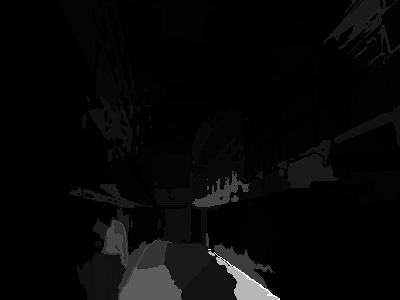}}&
\subfloat{\includegraphics[width=0.19\linewidth]{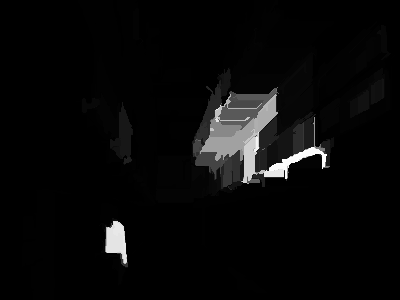}}\vspace{-0.15in}\\
\subfloat{\includegraphics[width=0.19\linewidth]{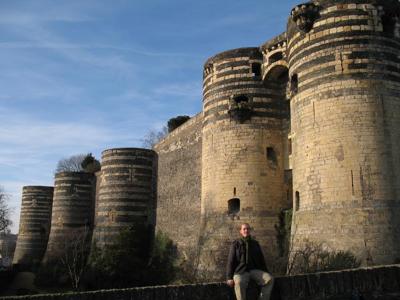}} &
\subfloat{\includegraphics[width=0.19\linewidth]{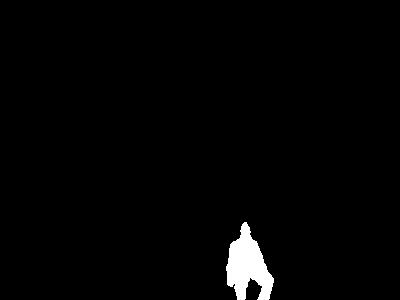}} &
\subfloat{\includegraphics[width=0.19\linewidth]{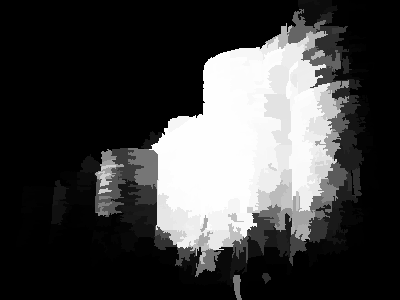}} &
\subfloat{\includegraphics[width=0.19\linewidth]{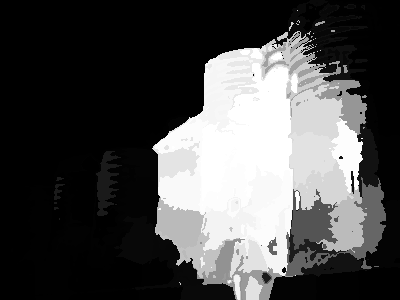}}&
\subfloat{\includegraphics[width=0.19\linewidth]{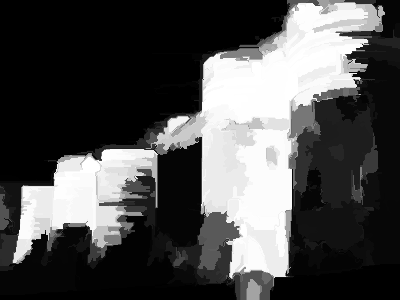}}\\
(a) & (b) & (c) & (d) & (e) \\
\end{tabular} \vspace{-0.1in}
\caption{Failure cases of our algorithm. (a) Input images, (b) Ground truths, Results of (c) our method, (d) MCDL~\cite{mcdl}, (e) MDF~\cite{mdf}.}\vspace{-0.15in}\label{fig:failure_cases}
\end{figure}

In \Figref{compare}, the PR-graph indicates our algorithm achieves the better performance than the previous works including MDF and MCDL which also utilize CNN models. Our algorithm shows the lowest MAE and the highest maximum F-measure score on most of the datasets. Visual comparisons of various methods are shown in \Figref{visualize_compare2}. We visualize the results from various difficult cases including low-contrast objects (row 1-3), complicate backgrounds (row 4-6), small salient objects (row 7-8), multiple salient objects(row 9-10) and touching boundary examples (row 11-12). Our algorithm shows especially good performance on images with low-contrast salient objects and complicated backgrounds, and also works well on other difficult scenes. 

In \Figref{failure_cases}, we reported some failure cases. The first and the second results contain correct salient objects but also highlight non-salient regions. The third and fourth examples have the extremely difficult scenes with a small, low-contrast and boundary touching the salient object. Because these kinds of data are not provided much by the training data, MSRA10K, we may further improve the performance with richer training data. For these difficult scenes, MCDL~\cite{mcdl} and MDF~\cite{mdf} also fail to find the salient objects precisely.

The running time of our algorithm was measured from the ECSSD dataset, where tested images were of size $400\times 300$. We used a server machine with intel i7 CPU, 8GB RAM and GTX Titan-Black for testing. Our model, developed by C++ and based on Caffe~\cite{caffe} library, took around 0.5 seconds per image. The training of our deep CNN took around 3 hours under the same environment. The short training time and testing time is also an advantage of our method. This is due to the sharing of our high level features which only need to be computed once for a whole image.

\section{Conclusion}
In this paper, we have introduced a new method to integrate the low-level and the high-level features for saliency detection. The Encoded Low-level Distance map (ELD-map) has stronger discriminative power than the original low-level feature distances to measure similarities or dissimilarities among superpixels. When concatenated with the high-level features from the deep CNN model (VGG16), our method shows the state-of-the-art performance in terms of both visual qualities and quantitative comparisons. As a future work, we are planning to explore more various CNN architectures to further improve the performance of our work.

\section*{Acknowledgement}
This work was partially supported by HRHRP(High Risk High Return Project of KAIST) and the MOTIE(The Ministry of Trade, industry \& Energy), Korea, under the Technology Innovation Program supervised by KEIT(Korea Evaluation Institute of Industrial Technology), 10045252, Development of robot task intelligence technology. Futhermore, this research was also supported by the National Research Foundation of Korea (NRF) under Grant NRF-2014R1A2A2A01003140.

{\small
\bibliographystyle{ieee}
\bibliography{egbib}

\begin{thebibliography}{10}\itemsep=-1pt

\bibitem{Achanta09cvpr}
R.~Achanta, S.~Hemami, F.~Estrada, and S.~Susstrunk.
\newblock Frequency-tuned salient region detection.
\newblock In {\em {Proc. of Computer Vision and Pattern Recognition (CVPR)}},
  2009.

\bibitem{slic}
R.~Achanta, A.~Shaji, K.~Smith, A.~Lucchi, P.~Fua, and S.~Susstrunk.
\newblock Slic superpixels compared to state-of-the-art superpixel methods.
\newblock {\em {IEEE Trans. Pattern Anal. Mach. Intell. (TPAMI)}},
  34(11):2274--2282, 2012.

\bibitem{appresize}
S.~Avidan and A.~Shamir.
\newblock Seam carving for content-aware image resizing.
\newblock 26(3):10, 2007.

\bibitem{SalObjSurvey}
A.~Borji, M.-M. Cheng, H.~Jiang, and J.~Li.
\newblock Salient object detection: A survey.
\newblock {\em ArXiv e-prints}, 2014.

\bibitem{SalObjBenchmark}
A.~Borji, M.-M. Cheng, H.~Jiang, and J.~Li.
\newblock Salient object detection: A benchmark.
\newblock {\em ArXiv e-prints}, 2015.

\bibitem{thur15k}
M.-M. Cheng, N.~Mitra, X.~Huang, and S.-M. Hu.
\newblock Salientshape: group saliency in image collections.
\newblock {\em The Visual Computer}, 30(4):443--453, 2014.

\bibitem{hc}
M.-M. Cheng, N.~J. Mitra, X.~Huang, P.~H.~S. Torr, and S.-M. Hu.
\newblock Global contrast based salient region detection.
\newblock {\em {IEEE Trans. Pattern Anal. Mach. Intell. (TPAMI)}},
  37(3):569--582, 2015.

\bibitem{pascal}
M.~Everingham, S.~A. Eslami, L.~Van~Gool, C.~K. Williams, J.~Winn, and
  A.~Zisserman.
\newblock The pascal visual object classes challenge: A retrospective.
\newblock {\em International Journal of Computer Vision}, 111(1):98--136, 2014.

\bibitem{Hariharan15cvpr}
B.~Hariharan, P.~Arbel{\'a}ez, R.~Girshick, and J.~Malik.
\newblock Hypercolumns for object segmentation and fine-grained localization.
\newblock In {\em {Proc. of Computer Vision and Pattern Recognition (CVPR)}},
  2015.

\bibitem{hornik1989multilayer}
K.~Hornik, M.~Stinchcombe, and H.~White.
\newblock Multilayer feedforward networks are universal approximators.
\newblock {\em Neural networks}, 2(5):359--366, 1989.

\bibitem{caffe}
Y.~Jia, E.~Shelhamer, J.~Donahue, S.~Karayev, J.~Long, R.~Girshick,
  S.~Guadarrama, and T.~Darrell.
\newblock Caffe: Convolutional architecture for fast feature embedding.
\newblock 2014.

\bibitem{mc}
B.~Jiang, L.~Zhang, H.~Lu, C.~Yang, and M.~Yang.
\newblock Saliency detection via dense and sparse reconstruction.
\newblock In {\em {Proc. of Int'l Conf. on Computer Vision (ICCV)}}, 2013.

\bibitem{drfi}
H.~Jiang, J.~Wang, Z.~Yuan, N.~Z. Y.~Wu, and S.~Li.
\newblock Salient object detection: A discriminative regional feature
  integration approach.
\newblock In {\em {Proc. of Computer Vision and Pattern Recognition (CVPR)}},
  2013.

\bibitem{Judd09iccv}
T.~Judd, K.~Ehinger, F.~Durand, and A.~Torralba.
\newblock Learning to predict where humans look.
\newblock In {\em {Proc. of Int'l Conf. on Computer Vision (ICCV)}}, 2009.

\bibitem{hdct}
J.~Kim, D.~Han, Y.~Tai, and J.~Kim.
\newblock Salient region detection via high-dimensional color transform.
\newblock In {\em {Proc. of Computer Vision and Pattern Recognition (CVPR)}},
  2014.

\bibitem{mdf}
G.~Li and Y.~Yu.
\newblock Visual saliency based on multiscale deep features.
\newblock In {\em {Proc. of Computer Vision and Pattern Recognition (CVPR)}},
  2015.

\bibitem{dsr}
X.~Li, H.~Lu, L.~Zhang, X.~Ruan, and M.~Yang.
\newblock Saliency detection via dense and sparse reconstruction.
\newblock In {\em {Proc. of Int'l Conf. on Computer Vision (ICCV)}}, 2013.

\bibitem{pascals}
Y.~Li, X.~Hou, C.~Koch, J.~M. Rehg, and A.~L. Yuille.
\newblock The secrets of salient object segmentation.
\newblock In {\em {Proc. of Computer Vision and Pattern Recognition (CVPR)}},
  pages 280--287. IEEE, 2014.

\bibitem{msra10k}
T.~Liu, J.~Sun, N.-N. Zheng, X.~Tang, and H.-Y. Shum.
\newblock Learning to detect a salient object.
\newblock In {\em {Proc. of Computer Vision and Pattern Recognition (CVPR)}},
  2007.

\bibitem{appdet}
P.~Luo, Y.~Tian, X.~Wang, and X.~Tang.
\newblock Switchable deep network for pedestrian detection.
\newblock In {\em {Proc. of Computer Vision and Pattern Recognition (CVPR)}},
  pages 899--906. IEEE, 2014.

\bibitem{mcdl}
W.~O. R.~Zhao, H.Li, and X.Wang.
\newblock Saliency detection by multi-context deep learning.
\newblock In {\em {Proc. of Computer Vision and Pattern Recognition (CVPR)}},
  2015.

\bibitem{appcrop}
C.~Rother, L.~Bordeaux, Y.~Hamadi, and A.~Blake.
\newblock Autocollage.
\newblock 25(3):847--852, 2006.

\bibitem{ILSVRC15}
O.~Russakovsky, J.~Deng, H.~Su, J.~Krause, S.~Satheesh, S.~Ma, Z.~Huang,
  A.~Karpathy, A.~Khosla, M.~Bernstein, A.~C. Berg, and L.~Fei-Fei.
\newblock {ImageNet Large Scale Visual Recognition Challenge}.
\newblock {\em International Journal of Computer Vision (IJCV)}, pages 1--42,
  April 2015.

\bibitem{appsum}
D.~Simakov, Y.~Caspi, E.~Shechtman, and M.~Irani.
\newblock Summarizing visual data using bidirectional similarity.
\newblock In {\em {Proc. of Computer Vision and Pattern Recognition (CVPR)}},
  pages 1--8. IEEE, 2008.

\bibitem{Simonyan14c}
K.~Simonyan and A.~Zisserman.
\newblock Very deep convolutional networks for large-scale image recognition.
\newblock {\em CoRR}, abs/1409.1556, 2014.

\bibitem{googlenet}
C.~Szegedy, W.~Liu, Y.~Jia, P.~Sermanet, S.~Reed, D.~Anguelov, D.~Erhan,
  V.~Vanhoucke, and A.~Rabinovich.
\newblock Going deeper with convolutions.
\newblock In {\em {Proc. of Computer Vision and Pattern Recognition (CVPR)}},
  2015.

\bibitem{legs}
L.~Wang, H.~Lu, X.~Ruan, and M.-H. Yang.
\newblock Deep networks for saliency detection via local estimation and global
  search.
\newblock In {\em {Proc. of Computer Vision and Pattern Recognition (CVPR)}},
  pages 3183--3192, 2015.

\bibitem{gabor}
T.~P. Weldon, W.~E. Higgins, and D.~F. Dunn.
\newblock Efficient gabor filter design for texture segmentation.
\newblock {\em Pattern Recognition}, 29(12):2005--2015, 1996.

\bibitem{hs}
Q.~Yan, L.~Xu, J.~Shi, and J.~Jia.
\newblock Hierarchical saliency detection.
\newblock In {\em {Proc. of Computer Vision and Pattern Recognition (CVPR)}},
  2013.

\bibitem{gmr}
C.~Yang, L.~Zhang, H.~Lu, X.~Ruan, and M.~Yang.
\newblock Saliency detection via graph-based manifold ranking.
\newblock In {\em {Proc. of Computer Vision and Pattern Recognition (CVPR)}},
  2013.

\bibitem{appreid}
R.~Zhao, W.~Ouyang, and X.~Wang.
\newblock Unsupervised salience learning for person re-identification.
\newblock In {\em {Proc. of Computer Vision and Pattern Recognition (CVPR)}},
  pages 3586--3593. IEEE, 2013.

\bibitem{rbd}
W.~Zhu, S.~Liang, Y.~Wei, and J.~Sun.
\newblock Saliency optimization from robust background detection.
\newblock In {\em {Proc. of Computer Vision and Pattern Recognition (CVPR)}},
  2014.

\end{thebibliography}
}

\end{document}